\documentclass[lettersize,journal]{IEEEtran}
\usepackage{changes}
\usepackage{amsmath,amsfonts}
\usepackage{algorithmic}
\usepackage{algorithm}
\usepackage{array}
\usepackage[caption=false,font=normalsize,labelfont=sf,textfont=sf]{subfig}
\usepackage[normalem]{ulem}
\useunder{\uline}{\ul}{}
\usepackage{textcomp}
\usepackage{stfloats}
\usepackage{url}
\usepackage{verbatim}
\usepackage{graphicx}
\usepackage{cite}
\usepackage{amssymb}
\usepackage{graphics} 
\usepackage{epsfig}
\usepackage{booktabs}
\usepackage{multirow}
\usepackage[utf8]{inputenc}
\usepackage{url}
\usepackage{booktabs}
\usepackage{amssymb}
\usepackage{bbding}
\usepackage{pifont}
\usepackage{wasysym}
\usepackage{utfsym}
\usepackage{fontawesome}
\usepackage{booktabs}
\usepackage{multirow}
\usepackage{amsmath}
\usepackage{hyperref}
\usepackage{mathrsfs}
\usepackage{txfonts}

\begin{document}

\title{ASCNet: Asymmetric Sampling Correction Network for Infrared Image Destriping}

\author{Shuai~Yuan,~Hanlin~Qin,~\IEEEmembership{Member,~IEEE},~Xiang~Yan,~\IEEEmembership{Member,~IEEE},~Shiqi~Yang,~Shuowen~Yang,~Naveed~Akhtar,~\IEEEmembership{Member,~IEEE},~Huixin~Zhou,~\IEEEmembership{Member,~IEEE}
\thanks{This work was supported in part by the Shaanxi Province Key Research and Development Plan Project under Grant 2022JBGS2-09, in part by the 111 Project under Grant B17035,
and in part by the Technology Area Foundation of China 2021-JJ-1244, 2021-JJ-0471, 2023-JJ-0148. \textit{(Corresponding authors:~Hanlin Qin, Xiang Yan.)}

Shuai~Yuan, Hanlin~Qin, Xiang~Yan, Shiqi~Yang and Shuaowen~Yang are with the School of Optoelectronic Engineering, Xidian University, Xi'an 710071, China. (email: yuansy@stu.xidian.edu.cn; hlqin@mail.xidian.edu.cn; xyan@xidian.edu.cn; 643722161@qq.com; shuowenyang@xidian.edu.cn)

Naveed Akhtar is with the School of Computing and Information Systems, Faculty of Engineering and IT, The University of Melbourne, Parkville VIC 3052, Australia (email: naveed.akhtar1@unimelb.edu.au).

Huixin Zhou is with the School of Physics, Xidian University, Xi’an 710071,
China (e-mail: hxzhou@mail.xidian.edu.cn).
}
}

\markboth{Journal of \LaTeX\ Class Files,~Vol.~14, No.~8, August~2021}%
{Shell \MakeLowercase{et al.}: A Sample Article Using IEEEtran.cls for IEEE Journals}

\maketitle
\begin{abstract}
\added{This is the pre-acceptance version, to read the final
version please go to IEEE TRANSACTION ON GEOSCIENCE
AND REMOTE SENSING on IEEE Xplore.}
In a real-world infrared imaging system, effectively learning a consistent stripe noise removal model is essential.
Most existing destriping methods cannot precisely reconstruct images due to cross-level semantic gaps and insufficient characterization of the global column features.
To tackle this problem, we propose a novel infrared image destriping method, called \textit{A}symmetric \textit{S}ampling \textit{C}orrection \textit{Net}work (ASCNet), that can effectively capture global column relationships and embed them into a U-shaped framework, providing comprehensive discriminative representation and seamless semantic connectivity.
Our ASCNet consists of three core elements: Residual Haar Discrete Wavelet Transform (RHDWT), Pixel Shuffle (PS), and Column Non-uniformity Correction Module (CNCM). 
Specifically, RHDWT is a novel downsampler that employs double-branch modeling to effectively integrate stripe-directional prior knowledge and data-driven semantic interaction to enrich the feature representation. 
Observing the semantic patterns crosstalk of stripe noise, PS is introduced as an upsampler to prevent excessive apriori decoding and performing semantic-bias-free image reconstruction. After each sampling, CNCM captures the column relationships in long-range dependencies. 
By incorporating column, spatial, and self-dependence information, CNCM well establishes a global context to distinguish stripes from the scene's vertical structures. 
Extensive experiments on synthetic data, real data, and infrared small target detection tasks demonstrate that the proposed method outperforms state-of-the-art single-image destriping methods both visually and quantitatively. Code is available at  \url{https://github.com/xdFai/ASCNet}.
\end{abstract}


\begin{figure}[t!]
    \centering
    \includegraphics[width=0.47\textwidth]{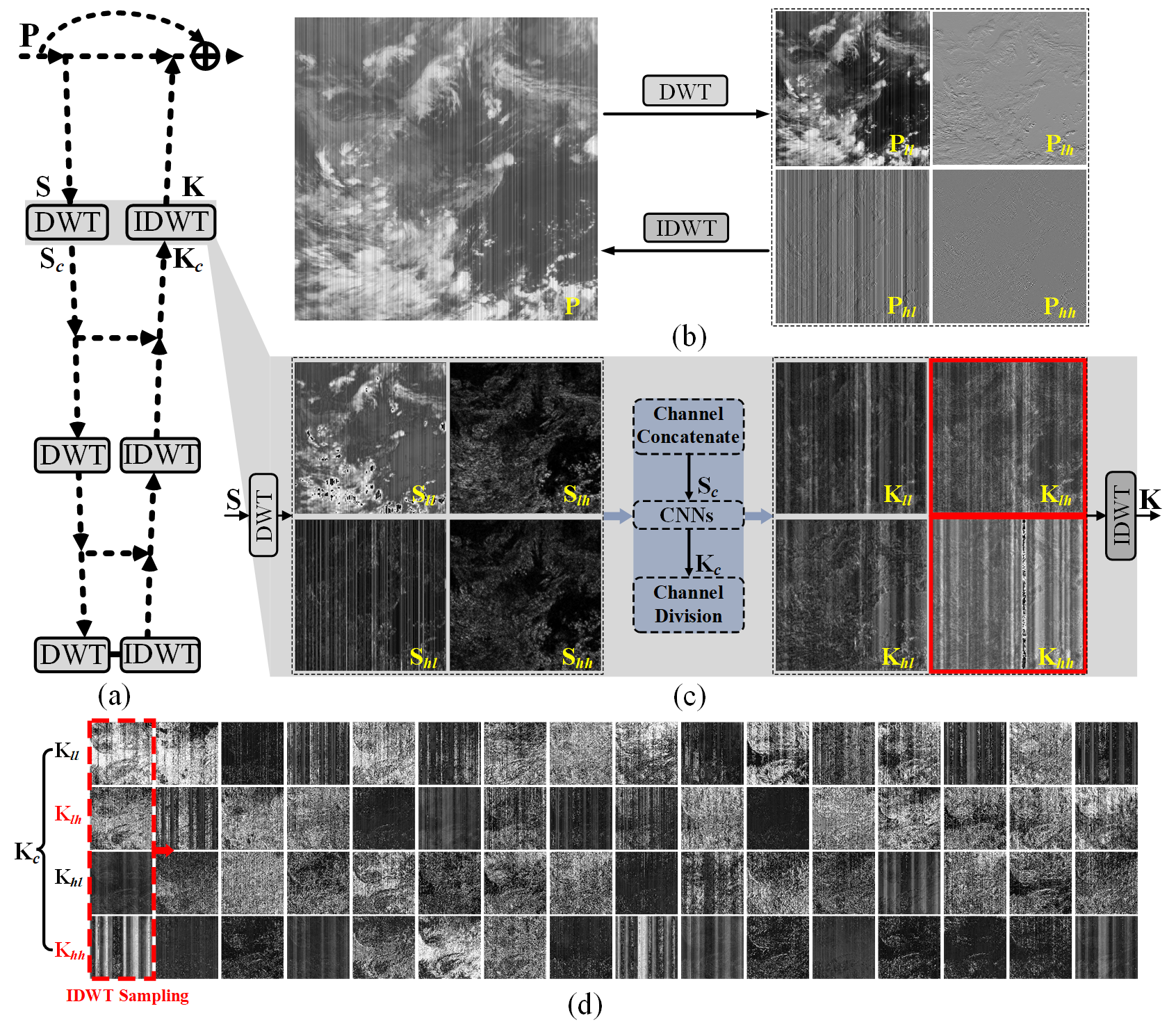}
    \vspace{-3mm}
    \caption{(a) Wavelet U-shaped neural network.
    (b) DWT decomposition and reconstruction of the stripy image $\mathbf{P}$.
    (c) DWT decomposition and reconstruction of the stripy feature $\mathbf{S}$ in wavelet U-shaped neural network.
    (d) Visualization of each channel of the crosstalk feature $\mathbf{K}_{c}$.
     }
    \label{fig:1}
    \vspace{-3mm}
\end{figure}

\section{Introduction}
\IEEEPARstart{I}{nfrared} (IR) imaging systems find wide applications in security monitoring, sea surveillance, and urban traffic~\cite{kou4},~\cite{23},~\cite{SCT}. 
However, due to the unique thermal imaging mechanism of infrared detectors, the varying responses of the Focal Plane Array (FPA), and its columnar signal readout manner, even corrected infrared images can still exhibit various types of column stripe noise under rapid external temperature fluctuations~\cite{41}.
This adversely affects the visual quality of images and downstream tasks, \emph{e.g.}, Infrared Small Target Detection (IRSTD)~\cite{72},~\cite{75}.
Thus, eradicating stripe noise from IR images is a crucial task.

A degraded IR stripy image ($\mathbf{I}_{D}$) can be represented as follows:
\begin{equation}
\mathbf{I}_{D}=\mathbf{I}_{N}+\mathbf{I}_{C}
\end{equation}
where $\mathbf{I}_{N}$ denotes the stripe noise and $\mathbf{I}_{C}$ represents the clean image. 
Recovering $\mathbf{I}_{C}$ from $\mathbf{I}_{D}$ is the primary objective of image destriping methods.
The challenges arising from the dynamic scenes have prompted the development of numerous techniques for single-image destriping~\cite{64},~\cite{31},~\cite{74}.
Early methods employed handcrafted features, such as filters~\cite{1},~\cite{4},~\cite{26}, data statistics~\cite{32},~\cite{66},~\cite{67}, and optimization techniques~\cite{76},~\cite{27},~\cite{92} to distinguish between background and stripe components.
Due to an over-reliance on empirical observation of image properties, these methods only achieve acceptable performance for relatively simple scenes.

\begin{figure}[t]
    \centering
    \includegraphics[width=0.48\textwidth]{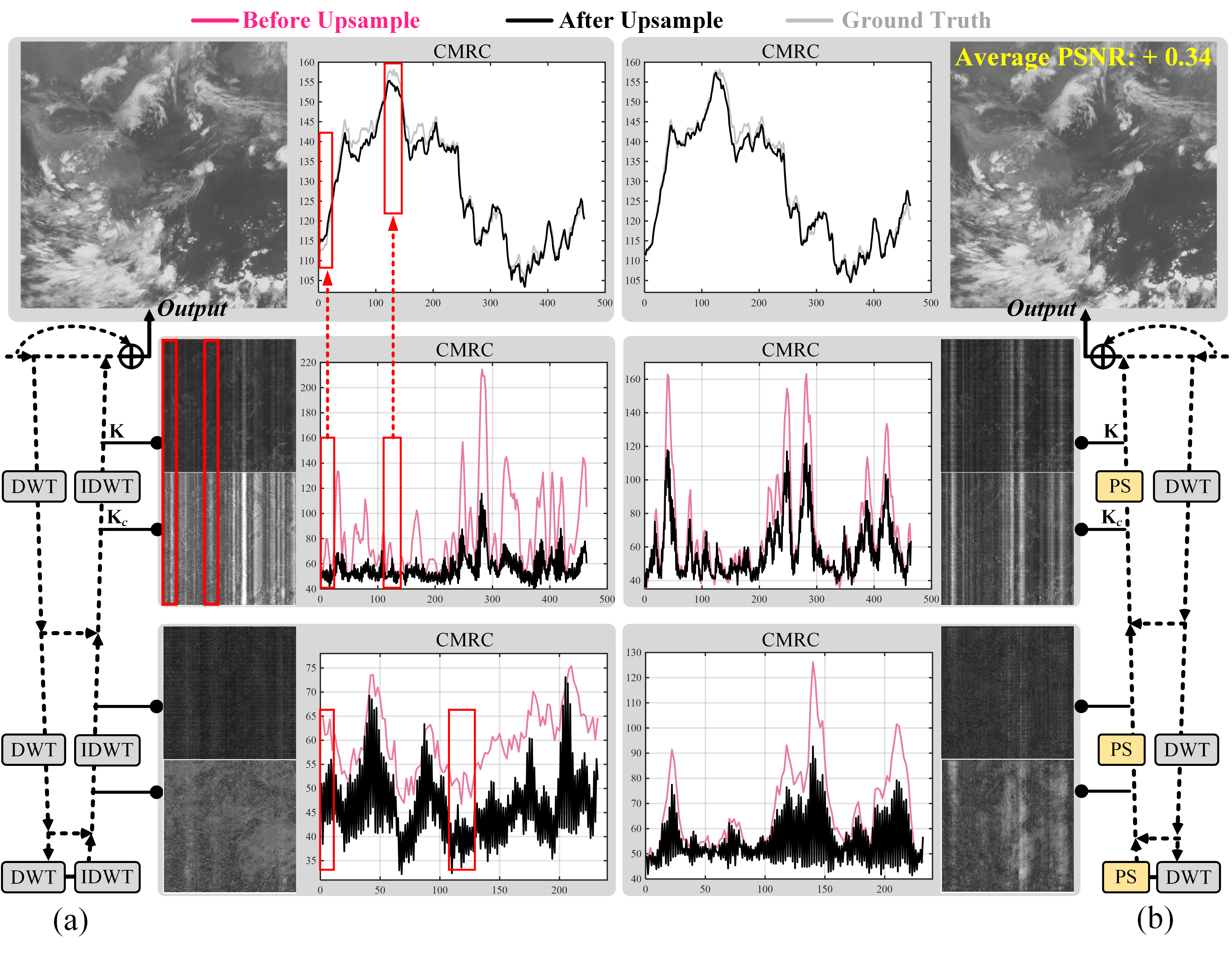}
    \vspace{-3mm}
    \caption{Insightful feature visualizations between (a) conventional \textbf{Symmetric Sampling}: DWT/IDWT, and (b) \textbf{Asymmetric Sampling}: DWT/PS during the last two stages of upsampling. 
    The Column-wise Mean Responding Curves (CMRC) of feature maps (maximum the channel-wise response) are provided to showcase the column semantic changes introduced by upsampling.
    The cross-level column semantic gap in symmetric sampling is highlighted in red boxes. We can observe that asymmetric sampling has stable semantic fluctuations and clearer pointers for noise reconstruction.}
    \vspace{-3mm}
    \label{fig:2}
\end{figure}

Owing to the powerful feature representation abilities of Convolution Neural Networks (CNNs), learning-based methods have demonstrated more promising destriping results for complex scenarios.
Initial approaches utilize shallow networks to mitigate stripe noise~\cite{34},~\cite{35}.
However, inadequate convolution layers lead to a small receptive field and limited high-level semantic understanding.
To address this issue,~\cite{6},~\cite{38}, and~\cite{36} employ residual learning strategies to achieve sufficient characterization with deeper networks.
Subsequently, multi-scale modeling~\cite{5},~\cite{39},~\cite{89} and attention mechanism~\cite{40},~\cite{88},~\cite{zilong1} have emerged as useful paradigms for IR image destriping. 
Considering the column signal readout of FPA, some methods also explored noise response characteristics to predict stripe distribution and achieved acceptable results~\cite{13},~\cite{11}.

Since stripe noise vertically compromises the image structure, \emph{i.e.}, it primarily degrades the image’s horizontal gradient. 
Benefiting from the powerful capabilities in gradient modeling, 2D Discrete Wavelet Transform (DWT) has received considerable attention in image destriping tasks~\cite{7},~\cite{37},~\cite{55}. 
As illustrated in Fig.~\ref{fig:1}(b), given a stripy image $\mathbf{P}$, DWT can reorganize stripe noise into the image's low-frequency sub-band $\mathbf{P}_{ll}$ and horizontal gradient sub-band $\mathbf{P}_{hl}$, while maintaining the cleanliness of the vertical gradient sub-band $\mathbf{P}_{lh}$ and diagonal sub-band $\mathbf{P}_{hh}$.
Subsequently, Inverse Discrete Wavelet Transform (IDWT) can reconstruct the image $\mathbf{P}$ without any loss based on the orthogonality of wavelet bases.
Motivated by the DWT's stripe decomposition ability, DWT and IDWT are naturally utilized as paired downsamplers and upsamplers in numerous U-shaped neural networks to build the internal directionality of stripe noise~\cite{8},~\cite{15},~\cite{DSCGAN}.
Despite the impressive performance achieved, these methods still struggle to precisely decouple the stripe noise and vertical background details.
We attribute this phenomenon to the following three challenges.
\vspace{-3mm}

\vspace{-1mm}
\subsection{IDWT disrupts the cross-level column semantic connectivity in wavelet U-shaped framework}
\label{Challenage one}
Given a stripe feature $\mathbf{S}$ before downsampling - see Fig.~\ref{fig:1}(c), although the DWT aggregates stripe noise into the $\mathbf{S}_{ll}$ and $\mathbf{S}_{hl}$, the channel-wise information interaction mechanism in convolution layers causes the stripe noise to repollute the vertical gradient sub-band $\mathbf{K}_{lh}$ and diagonal sub-band $\mathbf{K}_{hh}$.
With an inaccurate predetermined noise pattern distribution, the IDWT disrupts the column-wise response connectivity of features, creating a cross-level column semantic gap. 
As shown in Fig.~\ref{fig:2}(a), this semantic gap causes the restored image to deviate from the true column distribution.
\vspace{-2mm}

\subsection{Independent downsampling branch is insufficient to comprehensively depict stripy features}
Despite DWT modeling stripe noise directionality, it is limited to spatial sampling without channel interaction.
On the contrary, stride convolution considers spatial and full-semantic features but ignores the noise directional prior.
To sum up, an independent sampler often produces varying degrees of loss in semantics and structure, leading to insufficient feature representations~\cite{5},~\cite{8},~\cite{52}. 
\vspace{-2mm}

\subsection{Feature enhancement fails to highlight the  global column characteristic in long-range dependence}
An effective global column characteristic modeling is essential for the destriping task. While attention mechanisms~\cite{13},~\cite{DSCGAN},~\cite{Duan} and column calibration modules~\cite{12} have been employed, their effectiveness is limited by the absence of explicit modeling for column long-range dependencies, which hinders the exploration and modulation of characteristic column patterns within the global context.

\begin{figure*}[h]
    \centering
    \includegraphics[width=0.78\textwidth]{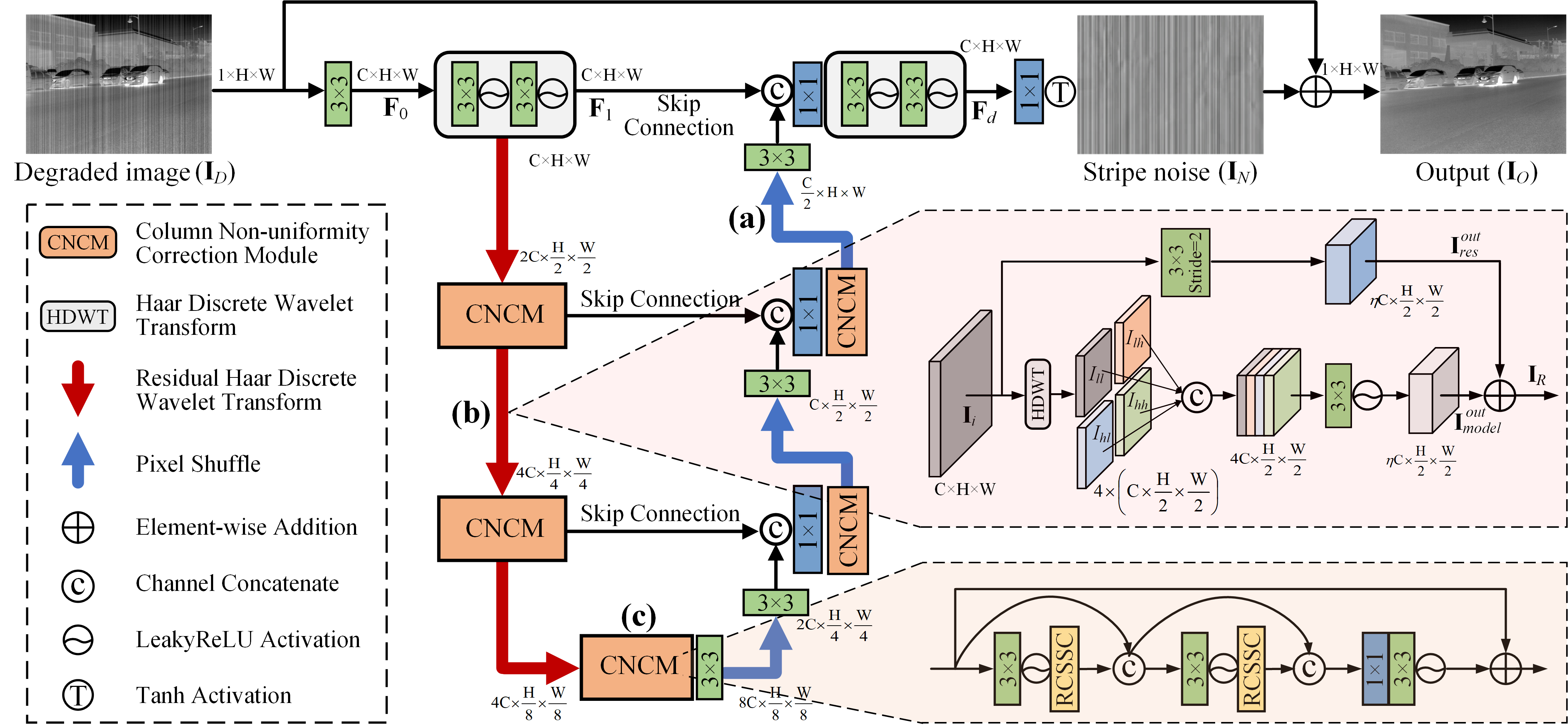}
    \vspace{-2mm}
    \caption{Overview of the proposed Asymmetric Sampling Correction Network (ASCNet). Three core modules of ASCNet are: (a) Pixel Shuffle (PS) to achieve semantic-bias-free image reconstruction, (b) Residual Haar Discrete Wavelet Transform (RHDWT) that enriches structural and semantic feature representation, and (c) Column Non-uniformity Correction Module (CNCM) nests the Residual Column Spatial Self-Correction (RCSSC) block to enhance column characteristics, spatial information, and long-range dependencies.}
    \vspace{-2mm}
    \label{fig:3}
    \vspace{-3mm}
\end{figure*}

To tackle these challenges, we propose a novel infrared image destriping method, called Asymmetric Sampling Correction Network (ASCNet), that can effectively construct a continuous semantic modeling and comprehensive representation of global column relationships.
As illustrated in Fig~\ref{fig:3}, ASCNet comprises three elements: 1) Pixel Shuffle (PS), which bridges the cross-level column semantic gap; 2) Residual Haar Discrete Wavelet Transform (RHDWT), for enriching feature characterization; and 3) the Column Non-uniformity Correction Module (CNCM), for globally enhancing columns.
The motivation of these components is as follows.


$\textbf{\textit{Challenge~A}}$: Considering that pixel shuffle directly reorganizes pixels without linear computation, its weaker prior assumption may better bridge the cross-level semantic gap~\cite{85}. 
Therefore, we build the symmetric sampling (DWT/IDWT) and asymmetric sampling (DWT/PS) in Fig.~\ref{fig:2} and showcase the feature maps and Column-wise Mean Responding Curves (CMRC) of the crosstalk feature $\mathbf{K}_{c}$ and reconstructed feature $\textbf{K}$.
It can be seen that the CMRC of asymmetric sampling is more stable, indicating pixel shuffle’s superior semantic articulation in stripy feature decoding. Therefore, pixel shuffle is utilized for ASCNet as upsamplers (Section~\ref{sec: MWCNN}).


$\textbf{\textit{Challenge~B}}$: 
Independently applying DWT for downsampling essentially serializes it with CNNs.
This blocks the flow of channel-wise semantic interaction.
Therefore, the proposed Residual Haar Discrete Wavelet Transform (RHDWT) has parallel model-driven and residual branches (Section~\ref{sec: RHDWT}).
Specifically, the model-driven branch utilizes Haar Discrete Wavelet Transform (HDWT) to incorporate stripe-directional prior knowledge to decompose the original feature.
The residual branch complements the information of the model-driven branch with data-driven cross-channel semantics.

$\textbf{\textit{Challenge~C}}$: Inspired by the success of the self-calibration strategy in image classification~\cite{46}, image deraining~\cite{77}, as well as the column-based correction module in remote sensing image destriping~\cite{12}, we present a Column Non-uniformity Correction Module (CNCM) to solve the last challenge (Section~\ref{sec: CNCM}).
Specifically, CNCM by nesting the Residual Column Spatial Self-Correction (RCSSC) block to achieve global column feature enhancement.
The RCSSC contributes three parts.
1) Column uniformity, which strengthens column characteristics to overcome the stripe noise column differences; 2) Spatial correlation, enhancing structural characterization of key areas to decouple stripe noise; 3) Self-dependence, establishing flexible remote dependencies that aggregate contextual information to fine-tune the global uniformity between different feature layers. 
Our main contributions are summarized below.

\begin{itemize}
    \item We propose an asymmetric sampling correction network for IR image destriping.
    In our technique, the stripe noise can be precisely decoupled from the scene by continuous column semantic modeling and comprehensive representations of global column relationships.
    
    \item  Based on the generally observed semantic crosstalk of stripe noise in wavelet U-shaped architectures, we employ pixel shuffle for the infrared image destriping to bridge the cross-level column semantic gap and restore images without semantic bias.
    
    \item We propose residual Haar discrete wavelet transform as a novel downsampler, which parallels stripe directional prior with semantic interaction for full feature description. 
    
    \item We devise a column non-uniformity correction module. The column characteristics of the global context can be fully captured by integrating long-range column uniformity, spatial correlation, and self-dependence.
\end{itemize}

\section{RELATED WORK}

In this section, we briefly review the major works in single infrared image destriping.
Following that, we introduce the studies on wavelet U-shaped structures in image processing.

\vspace{-3mm}
\subsection{Single Infrared Image Destriping}
\label{Sec: II-A}
\subsubsection{Model-Based Methods}
Traditional paradigms rely on the structural properties observed in the stripe and background. 
Cao et al.~\cite{1} used two orthogonal filters, one-dimensional row-guided filter and column-guided filter, to remove the stripe noise in a phased manner.
Yet, purely handcrafted filters inevitably blur the image texture.
Since the grayscale response of each column in the image exhibits a similar statistical distribution, Tendero et al.~\cite{32} employed midway histogram equalization to fine-tune pixel intensities within columns.
However, the distribution of real scenes is quite complex, and the statistical information generated by each detector often exhibits high variability.
To solve this problem, Chang et al.~\cite{3} combined unidirectional total variation and sparse representation to propose a joint model for removing random stripe noise. Similarly, Li et al.~\cite{33} considered the gradient and global sparsity of the stripe noises, treating the components along the stripe direction as a fidelity term.

\subsubsection{Learning-Based Methods} 
This category of methods mainly depends on convolutional neural networks (CNNs). 
For instance, Xiao et al.~\cite{35} integrated local and global information into CNNs and designed a network to optimize the edge preservation performance.
He et al.~\cite{6} proposed the Deep Learning-based Stripe Non-Uniformity Correction (DLS-NUC) model, utilizing an end-to-end residual deep network. This work also introduces the first comprehensive cubic noise simulation model.
Recently, multi-scale strategies have been widely used due to their ability to extract features from different receptive fields. \emph{e.g.}, Xu et al.~\cite{39} introduced a multi-scale dense connection structure that can extract fine and coarse features for clean image recovery.
Similarly, multi-level encoder and decoder structures also get much attention.
Chang et al.~\cite{5} used a U-shape sampling network to integrate multi-scale and long–short-term residual information for correcting image non-uniformity.
Unlike these multi-level U-shaped frameworks, which utilize existing scale transform structures, we devise a novel RHDWT downsampler to enrich the feature representation.

Recently, several advanced methods have combined noise response or structural characteristics with attention mechanisms to achieve better results~\cite{zilong2}.
Guan et al.~\cite{13} designed a cascade residual attention CNN model that estimates the gain and bias of the noise response using two concatenated sub-networks.
Taking into account the signal readout by column, Li et al.~\cite{12} designed a Column-Spatial Correction Network (CSCNet), utilizing CCMs with spatial attention to ensure local consistency in uniform regions and global uniformity in the image.
Although satisfactory results are achieved, these methods fail to fully exploit the characteristic column patterns in a global context due to a lack of long-range dependence.
Different from CCM~\cite{12}, our CNCM calculates the average and max response of the columns in two steps. Moreover, column responses are expanded at the final stage to vertically calibrate the original features. Finally, a self-calibration module is incorporated to more flexibly aggregate column global contextual information.

\vspace{-3mm}
\subsection{Wavelet U-shaped Structures in Image Processing}
\label{Sec: II-B}
Owing to the discrete wavelet transform can extract frequency information and acquire an acceptable receptive field~\cite{43}, a large amount of wavelet U-shaped methods have been proposed for image processing.
\emph{e.g.}, Wu et al.~\cite{91} present a self-attention memory-augmented wavelet-CNN for anomaly detection, in which DWT and IDWT are paired as samplers.
Liu et al.~\cite{59} proposed the Multi-level Wavelet Convolutional Neural Networks (MWCNN) for image denoising. Notable, MWCNN is a typical representative of the wavelet U-shaped structures.
As mentioned, stripe noise primarily disrupts the image's horizontal gradient. DWT gives a clearer semantic in the destriping task: Accurately aggregates noise into low frequency and horizontal gradient sub-bands.
Therefore, Guan et al.~\cite{7} first proposed a Wavelet Deep Neural Network for Stripe Noise Removal (SNRWDNN). This network was designed to directly learn the mapping between wavelet sub-bands and utilize inverse discrete wavelet transform to reconstruct the clean image.
Inspired by MWCNN, Chang et al.~\cite{8} presented a Two-Stream Wavelet Enhanced U-Net (TSWEU) for stripe noise estimation.
Most recently, DSCGAN~\cite{DSCGAN} introduced an unsupervised multi-level wavelet structure as a destriping generator to capture the real sense stripe distribution.

Despite the wavelet transforms providing CNNs with the directional characteristics of stripe noise, the wavelet upsampler with a predetermined noise semantic distribution will disrupt the semantic connectivity of the feature decoding.
Unlike previous wavelet U-shaped structures for destriping, we use the pixel shuffle as a semantic-bias-free upsampler for stable image restoration.

\section{PROPOSED METHOD}
We first briefly introduce the DWT in a U-shaped framework in Section~\ref{sec: MWCNN}, and follow by presenting the overall network architecture of ASCNet in Section~\ref{sec:ONA}. Then, we present the technical details of the RHDWT and CNCM in Section~\ref{sec: RHDWT} and Section~\ref{sec: CNCM}, respectively.

\subsection{DWT for Stripy Features in U-shaped Framework}
\label{sec: MWCNN}
DWT for image processing is extensively discussed in MWCNN~\cite{59}.
Conversely, we concentrate on the advantages and limitations of DWT in handling stripe noise features.
As shown in Fig.~\ref{fig:1}(c), given an input stripy feature $\mathbf{S} \in \mathbb{R}^{C \times H \times W}$, 2D DWT is utilized with four filters: a low-pass filter $\boldsymbol{{\lambda}}_{ll}\in \mathbb{R}^{2 \times 2}$, and three high-pass filters $\boldsymbol{{\lambda}}_{lh} \in \mathbb{R}^{2 \times 2}$, $\boldsymbol{{\lambda}}_{hl} \in \mathbb{R}^{2 \times 2}$, $\boldsymbol{{\lambda}}_{hh} \in \mathbb{R}^{2 \times 2}$ to decompose $\mathbf{S}$ into four sub-bands:
low-frequency sub-band
$\mathbf{S}_{ll}  \in  \mathbb{R}^{C \times H/2 \times W/2}$,
vertical gradient sub-band
$\mathbf{S}_{lh}  \in  \mathbb{R}^{C \times H/2 \times W/2}$,
horizontal gradient sub-band
$\mathbf{S}_{hl}  \in  \mathbb{R}^{C \times H/2 \times W/2}$,
and diagonal sub-band
$\mathbf{S}_{hh}  \in  \mathbb{R}^{C \times H/2 \times W/2}$.
Without prejudice to generality, using the Haar wavelet as an illustration, the four filters are defined as follows:
\begin{equation}
\begin{aligned}
\label{equ: 2}
\boldsymbol{{\lambda}}_{ll} & =\left[\begin{array}{cc}
1 & 1 \\
1 & 1
\end{array}\right],
\quad\quad\boldsymbol{{\lambda}}_{lh}&=\left[\begin{array}{cc}
-1 & -1 \\
1 & 1
\end{array}\right], \\
\boldsymbol{{\lambda}}_{hl} & =\left[\begin{array}{cc}
-1 & 1 \\
-1 & 1
\end{array}\right], 
\quad~\boldsymbol{{\lambda}}_{hh}&\!=\left[\begin{array}{cc}
1 & -1 \\
-1 & 1
\end{array}\right].
\end{aligned}
\end{equation}
The operation of HDWT for feature $\mathbf{S}$ is defined as:

\begin{align}
& {\mathbf{S}_{a}}=\boldsymbol{\lambda}_{a}\otimes{\mathbf{S}},~a \in \left\{ll, lh, hl, hh\right\},
\label{equ: 1}
\end{align}
where $\otimes$ represents the convolution operator with stride set to 2.
Due to the dimensional mismatch between the filters ${{\lambda}_{a}}$ and the stripy feature $\mathbf{S}$, broadcasting is applied in Eq.~\ref{equ: 1}.
Under this computational approach, the HDWT differs from convolution layers in two aspects:
1) Wavelet filters have constant parameters, allowing for the incorporation of prior information in extracting the directional features of stripe noise;
2) Each wavelet filter depth-wise calculates every channel of $\mathbf{S}$. In other words, HDWT only extracts spatial features without semantic interaction.
Therefore, HDWT can aggregate stripe noise into the low-frequency sub-band ($\mathbf{S}_{ll}$) and horizontal gradient sub-band ($\mathbf{S}_{hl}$) in Fig.~\ref{fig:1}(c).

In muti-level wavelet CNNs, the four wavelet sub-bands are next channel-wise concatenated to acquire fusion feature $\mathbf{S}_{c} \in \mathbb{R}^{4C \times H/2 \times W/2}$, followed by a series of convolution layers ${f_{s}(\cdot)}$ to get the high-level feature $\mathbf{K}_{c} \in \mathbb{R}^{4C \times H/2 \times W/2}$.
\begin{align}
& \mathbf{S}_{c} = [\mathbf{S}_{ll}, \mathbf{S}_{lh},\mathbf{S}_{hl} ,\mathbf{S}_{hh}], \\
& \mathbf{K}_{c} = {f_{s}}(\mathbf{S}_{c}).
\end{align}
where the $[\cdot]$ is the channel-wise concatenation,
Take note that the weights of conventional convolution layers are learnable, and the results of each step of spatial feature computation will be summed along the channel dimensions to encapsulate cross-semantic information.
Therefore, every channel of $\mathbf{K}_{c}$ is filled with stripe noise in Fig.~\ref{fig:1}(d).

We next perform an inverse wavelet transform on the high-level feature $\mathbf{K}_{c}$ for upsampling. 
First, $\mathbf{K}_{c}$ is split into four equal branches as follows:
\begin{equation}
[\mathbf{K}_{1}, \mathbf{K}_{2},\mathbf{K}_{3} ,\mathbf{K}_{4}] = {{Chunk}_{4}}(\mathbf{K}_{c}),
\end{equation}
where ${{Chunk}_{4}}(\cdot)$ denotes dividing the feature vector into four equal parts along the channel dimension, $\mathbf{K}_{i} \in \mathbb{R}^{C \times H/2 \times W/2}~(i \in \left\{1,2,3,4\right\})$ represent the four group of divided features.
When performing invert transform, $\mathbf{K}_{i}~(i \in \left\{1,2,3,4\right\})$ can be renamed as four wavelet sub-bands $\mathbf{K}_{a}~(a \in \left\{ll, lh, hl, hh\right\})$,
this operation is thus defined as follows:
\begin{align}
& \mathbf{K}^{k} =\Psi(\mathbf{K}_{ll}^{k}, \mathbf{K}_{lh}^{k},\mathbf{K}_{hl}^{k} ,\mathbf{K}_{hh}^{k}), \\
& \mathbf{K} = [\mathbf{K}^{1}, \mathbf{K}^{2} ,\dots ,\mathbf{K}^{c}],
\end{align}
where $\Psi(\cdot)$ refers to the Invert Haar Discrete Wavelet Transform (IHDWT) operator, which samples each channel as shown in Fig.~\ref{fig:1}(d).
Afterward, four linear operations are applied to the wavelet sub-bands using the inverse matrix in Eq.~\ref{equ: 2} to reconstruct each channel feature, as described in~\cite{59}.
Next, we reveal the shortcomings of IHDWT for upsampling stripe features.
First, employing IHDWT means treating $\mathbf{K}_{i}~(i \in \left\{1,2,3,4\right\})$ as the wavelet sub-bands.
However, the wavelet vertical gradient sub-band $\mathbf{K}_{lh}$ and diagonal sub-band $\mathbf{K}_{hh}$ are contaminated. 
This contradicts the original intention of discrete wavelet transform: leveraging the directional prior of stripes, resulting in a cross-level column semantic gap in Fig~\ref{fig:2}(a).
In contrast, pixel shuffle directly reorganizes features without linear calculation. 
Hence, it exhibits stronger column coherence in Fig~\ref{fig:2}(b).

\begin{figure*}[t]
    \centering
    \includegraphics[width=0.85\textwidth]{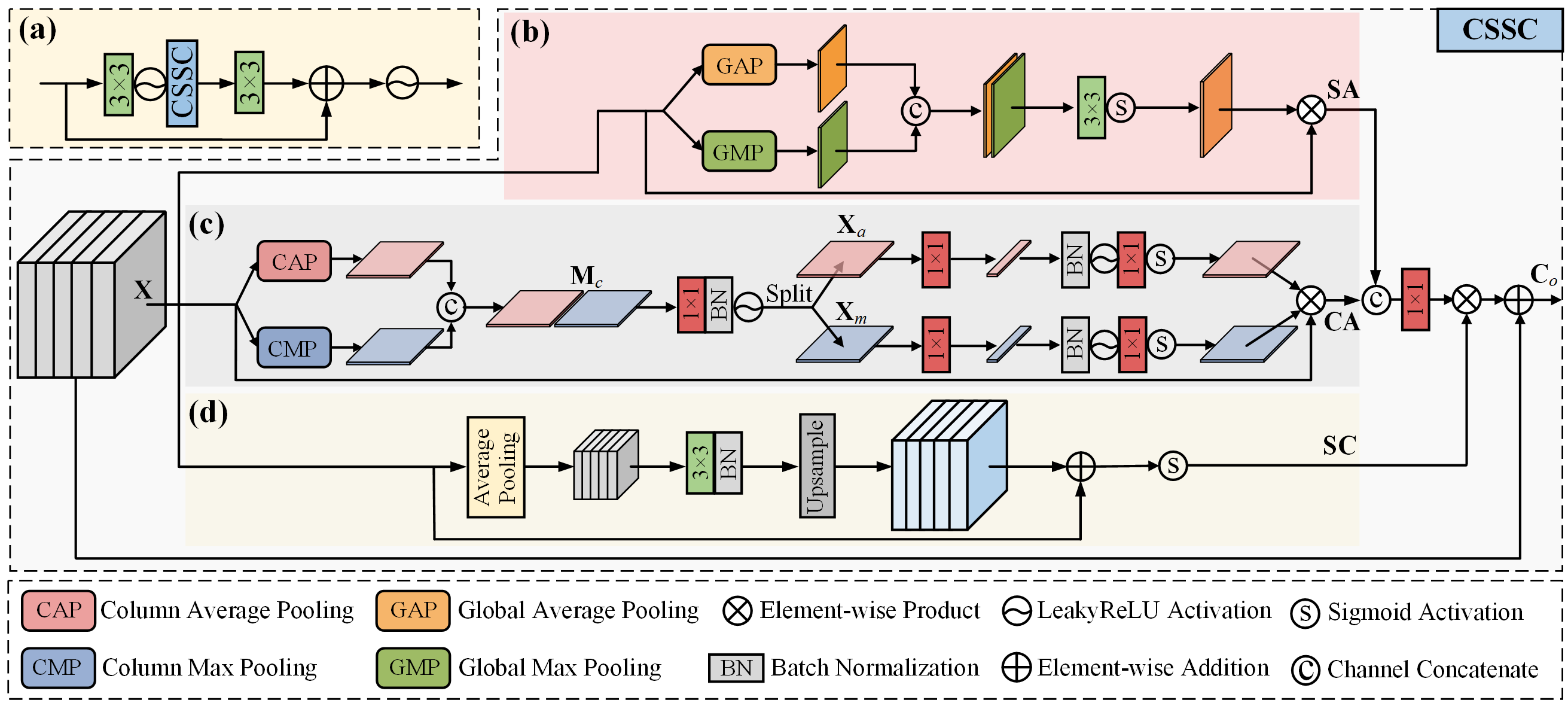}
    \vspace{-2mm}
    \caption{Architecture of the Residual Column Spatial Self-Correction (RCSSC) block. (a) RCSSC incorporates residual connection into the Column Spatial Self-Correction (CSSC) block. 
    The core modules of CSSC are: (b) Spatial Attention Branch (SAB) that utilizes spatial correlation to enhance structural characterization of key areas, (c) Column Attention Branch (CAB) that strengthens column characteristics to eliminate the stripe noise column differences and (d) Self-Calibrated Branch (SCB) that build remote dependencies to fine-tune the global uniformity.}
    \label{fig:4}
    \vspace{-3mm}
\end{figure*}

\subsection{Overall Network Architecture}
\label{sec:ONA}
The overall architecture of ASCNet is illustrated in Fig.~\ref{fig:3}.
It incorporates Column Non-uniformity Correction Modules (CNCM) into an asymmetric U-shaped structure to learn the residual mapping from $\mathbf{I}_{D}$ to the clean image $\mathbf{I}_{C}$.
Specifically, given a degraded image $\mathbf{I}_{D} \in \mathbb{R}^{1 \times H \times W}$, ASCNet applies a convolution to $\mathbf{I}_{D}$ to obtain low-level feature ${\mathbf{F}_{0}} \in \mathbb{R}^{C \times H \times W}$. 
Then, ${\mathbf{F}_{0}}$ is further enhanced through two consecutive convolution and LeakyReLU units to obtain shallow feature $\mathbf{F}_{1}\in \mathbb{R}^{C \times H \times W}$.
Subsequently, this shallow feature is sequentially downsampled and feature-enhanced by passing through three groups of RHDWT and CNCM to achieve information encoding.
Based on the analysis in Section~\ref{sec: MWCNN}, pixel shuffles are utilized as upsamplers in the decoding stage. 
To capture low-level details and high-level semantics, long-skip connections are adopted to fuse encoded and decoded features, and $3\times3$ convolutions are applied to ensure decoded features have the same channel dimension with encoded features.
Subsequently, the $1\times1$ convolution is performed to reduce channels of fused features by half.
CNCM is employed to enhance the fused features, which finely split stripe noise and capture textural details. 
Finally, further two consecutive convolution and LeakyReLU units are utilized to enrich high-resolution features $\mathbf{F}_{d}$.
A $1\times1$ convolution and Tanh activation are applied to $\mathbf{F}_{d}$ to generate the residual image $\mathbf{I}_{N} \in \mathbb{R}^{1 \times H \times W}$, which is then added with the degraded image $\mathbf{I}_{D} \in \mathbb{R}^{1 \times H \times W}$ to obtain the restored image $\mathbf{I}_{O} \in \mathbb{R}^{1 \times H \times W}$. 
We will discuss the effectiveness of pixel shuffle in Section~\ref{sec: Asymmetric}.

\vspace{-3mm}
\subsection{Residual Haar Discrete Wavelet Transform}
\label{sec: RHDWT}
The independent sampler often produces varying degrees of semantic and structural information loss~\cite{5},~\cite{8}.
Consequently, we propose an RHDWT. It integrates a model-driven branch with a residual branch to address this issue.
As shown in Fig.~\ref{fig:3}(b), given an input feature $\mathbf{I}_{i} \in \mathbb{R}^{C \times H \times W}$, HDWT is first utilized to decompose $\mathbf{I}_{i}$ into four subgraphs and concatenate them.
Subsequently, 3 $\times$ 3 convolution and LeakyReLU activation are employed to squeeze the concatenated sub-bands to get the model-driven branch's output $ \mathbf{I}^{out}_{model} \in \mathbb{R}^{\eta C \times H/2 \times W/2}$, where $\eta$ denotes the channel expansion factor.
In the residual branch, a 3 $\times$ 3 convolution with a step size two is used to aggregate spatial and semantic features to get output $\mathbf{I}^{out}_{res} \in \mathbb{R}^{\eta C \times H/2 \times W/2}$.
Eventually, we add the outputs of two branches to get the result $\mathbf{I}_{R} \in \mathbb{R}^{\eta C \times H/2 \times W/2}$ as follows:
\begin{equation}
\begin{aligned}
\mathbf{I}_{R}~\!&=\mathbf{I}^{out}_{model} + \mathbf{I}^{out}_{res},\\
&= f^{\delta}_{3\times3}([\Phi(\mathbf{I}_{i})])+ {f^{s=2}_{3\times3}}(\mathbf{I}_{i}),
\end{aligned}
\end{equation}
in which $\Phi(\cdot)$ denotes the HDWT operation, ${f^{\delta}_{3\times3}}$ represents 3 $\times$ 3 convolution and LeakyReLU activation, and ${f^{s =2}_{3\times3}}$ is 3 $\times$ 3 convolution with step size set 2.
We will discuss the validity of RHDWT in Section~\ref{sec: RHDWT_Ablation}.

\vspace{-2mm}
\subsection{Column Non-uniformity Correction Module}
\label{sec: CNCM}
\vspace{-1mm}
The Column Non-uniformity Correction Module (CNCM) is depicted in Fig.~\ref{fig:3}(c), which nests the RCSSC block into a Densely Connected Residual (DCR) structure~\cite{17}. DCR can induce feature reusing and enhance information flow~\cite{45}. 
Fig. \ref{fig:4}(a) illustrates RCSSC, a structure that integrates residual connections into the CSSC block.
The CSSC block comprises three branches: the Spatial Attention Branch (SAB), the Column Attention Branch (CAB), and the Self-Calibrated Branch (SCB). 
Inspired by the effective self-calibrated dual attention mechanism blocks for image deraining~\cite{77}, given an input feature $\mathbf{X}  \in \mathbb{R}^{C \times H \times W}$, our CSSC process is defined as:
\begin{equation}
\begin{aligned}
\label{equ:6}
\mathbf{C}_{o}~\!&=f_{1\times1}({[SAB(\mathbf{X}),CAB(\mathbf{X})]})\odot{SCB(\mathbf{X})}+\mathbf{X}\\
&=f_{1\times1}({[\mathbf{SA},\mathbf{CA}]})\odot{\mathbf{SC}}+\mathbf{X},
\end{aligned}
\end{equation}
in which $\mathbf{C}_{o} \in \mathbb{R}^{C \times H \times W}$ is the output of CSSC, $f_{1\times1}$ denotes 1×1 convolution, $\odot$ is the broadcasted Hadamard product.
$\mathbf{SA}$, $\mathbf{CA}$, and $\mathbf{SC}$ are the output of the module SAB, CAB and SCB, respectively.

\subsubsection{Spatial Attention Branch}
SAB is employed to enhance the spatial correlation~\cite{71}.
As illustrated in Fig.~\ref{fig:4}(b), given an input feature $\mathbf{X} \in \mathbb{R}^{C \times H \times W}$, global average and max pooling along the channel dimension are applied to capture the different semantic responses. Then, these responses are concatenated and aggregated to acquire spatial weights, followed by multiplying with the input feature $\mathbf{X}$ to generate output feature $\mathbf{SA} \in \mathbb{R}^{C \times H \times W}$.
Mathematically,
\begin{equation}
\mathbf{SA}=f^{s}_{3\times3}([{f_{spatial}^{Avg}(\mathbf{X}),f_{spatial}^{Max}(\mathbf{X})}])\odot{\mathbf{X}},
\end{equation}
where $f^{s}_{3\times3}$ represents 3 $\times$ 3 convolution and Sigmoid activation, ${f_{spatial}^{Avg}(\cdot)}$ and ${f_{spatial}^{Max}(\cdot)}$ denotes the channel-wise average and max pooling.

\subsubsection{Column Attention Branch}
Since the same amplifier is shared by column pixels, features within the same column should be assigned similar correction coefficients to emphasize column dependencies.
Considering global pooling can capture richer high-level information~\cite{70}, in Fig.~\ref{fig:4}, we apply the column average pooling ${f_{column}^{Avg}}(\cdot)$ and max pooling ${f_{column}^{Max}}(\cdot)$ with pooling kernels size (H,1) to each column of the input feature $\mathbf{X} \in \mathbb{R}^{C \times H \times W}$.
Since the different pooling responses correspond to the same columns, they are 
concatenated along the channel dimension to capture this shared characteristic as follows:
\begin{equation}
\mathbf{M}_{c}=[{f_{column}^{Avg}(\mathbf{X}),f_{column}^{Max}(\mathbf{X})}].
\end{equation}
The concatenated feature $\mathbf{M}_{c} \in \mathbb{R}^{2C \times 1 \times W}$ 
is shared to achieve the information interacted and split into two equal branches as follows:
\begin{equation}
\mathbf{X}_{a},\mathbf{X}_{m} = {{Chunk}_{2}}(\mathcal{CBL}(\mathbf{M}_{c})),
\end{equation}
where ${{Chunk}_{2}}(\cdot)$ denotes channel-wise dividing the feature vector into two equal parts, $\mathcal{CBL}(\cdot)$ represents a set of Conv1$\times$1+BN+LeakyReLU functions.
Then, the attention models are employed to obtain the column weights.
The column attention matrix $\mathbf{CA} \in \mathbb{R}^{C \times H \times W}$ can be written as:
\begin{equation}
 \mathbf{CA} = \mathbf{X}\odot{\mathcal{F}_{column}^{Avg}}(\mathbf{X}_{a})\odot{\mathcal{F}_{column}^{Max}}(\mathbf{X}_{m}),
\end{equation}
in which ${\mathcal{F}_{column}^{Avg}}(\cdot)$ and ${\mathcal{F}_{column}^{Max}(\cdot)}$ represent a series of Conv1$\times$1 +BN+ReLU+Conv1$\times$1+Sigmoid functions for channel attention mechanisms~\cite{18}.

\begin{table*}[t]
\centering
\caption{The average PSNR and SSIM of various destriping methods on three test sets with different types of stripe noise. The best and the second-best results are highlighted in bold and underlined, respectively.}
\label{tab1}
\vspace{-1mm} 
\renewcommand{\arraystretch}{1.2}
\setlength\tabcolsep{1.35mm}
\begin{tabular}{ccccccccccccccc}
\hline  \hline
                Category       & Index     & Degraded & GF   & UTV  & LRSID  & SEID  & DMRN    & DLS-NUC & SNRWDNN & TSWEU  & Restormer  & DSCGAN  & ASCNet  \\ \hline
       \multicolumn{14}{c}{ICSNR}    \\ \hline         
 \multirow{2}{*}{Gaussian}      & PSNR$\uparrow$  & 24.80  & 34.62 & 33.27 & 34.92 & 33.09 & 32.61 & 33.12 & 35.37 & 36.76 & 37.11 & \underline{38.82}  & \textbf{40.26} \\ 
                                & SSIM$\uparrow$  & 0.5742 & 0.9772 & 0.9673 & 0.9833 & 0.9583 & 0.9822  & 0.9615  &  0.9841 & 0.9900 & 0.9888 & \underline{0.9911} & \textbf{0.9952}  \\
 \multirow{2}{*}{Uniform}    & PSNR$\uparrow$     & 27.06  & 36.47 & 34.58 & 36.46 & 33.87 & 33.09 & 35.93 & 36.88 & 37.65 & 38.30 & \underline{40.58} & \textbf{41.53} \\
 & SSIM$\uparrow$  & 0.6612   & 0.9843  & 0.9759  &  0.9876  & 0.9655  & 0.9869  & 0.9755  & 0.9875  & 0.9907 & 0.9902 & \underline{0.9933}  & \textbf{0.9959}  \\
 \multirow{2}{*}{Periodical} & PSNR$\uparrow$  & 26.94  & 35.32 & 33.42 & 34.52 & 32.66 & 31.84 & 33.98 & 35.17 & 35.28 & 34.51 &  \underline{36.73}  & \textbf{37.01} \\
 & SSIM$\uparrow$  & 0.5774   & 0.9897  & 0.9783  &  0.9903 & 0.9692 & 0.9892  & 0.9712  & 0.9895  & 0.9906 & 0.9891 &  \underline{0.9926}  & \textbf{0.9952}  \\  
  \multirow{2}{*}{Mixed}      & PSNR$\uparrow$  & 26.13  & 32.61 & 31.88 & 33.13 & 32.97 & 30.80 & 32.96 & 32.77 & 34.53  & \underline{35.56} & 34.22  & \textbf{36.50} \\
 & SSIM$\uparrow$  & 0.6160   & 0.8743  & 0.8733  & 0.9011  & \underline{0.9398}  & 0.8763  & 0.9064  & 0.8750  & 0.9208  & 0.9120 & 0.8980   & \textbf{0.9589}  \\ \hline
\multicolumn{14}{c}{INFRARED}    \\ \hline   
 \multirow{2}{*}{Gaussian}   & PSNR$\uparrow$  & 23.94  & 33.95 & 33.23 & 35.04   & 33.83  & 31.48 & 33.45 &  36.23 & 37.44  & 37.36 & \underline{39.14}  & \textbf{40.10} \\
                             & SSIM$\uparrow$  & 0.3780  & 0.9520  & 0.9388  & 0.9614  & 0.8995 & 0.9502  & 0.9302  & 0.9676 & 0.9818  & 0.9792 & \underline{0.9809}  & \textbf{0.9889}  \\
 \multirow{2}{*}{Uniform}    & PSNR$\uparrow$  & 27.12  & 36.86 & 35.71 & 37.43 & 38.12 & 32.43 & 39.25 & 38.67 & 39.00 & 39.09 & \underline{41.67}  & \textbf{42.10} \\
                             & SSIM$\uparrow$  & 0.5026   & 0.9711  & 0.9622  & 0.9695 & 0.9420 & 0.9669  & 0.9687  & 0.9763 & 0.9818 & 0.9819 & \underline{0.9844}  & \textbf{0.9905}  \\
 \multirow{2}{*}{Periodical} & PSNR$\uparrow$  & 26.43  & 35.80 & 34.12 & 34.36 &  35.48 & 31.013 & 35.94 & 35.31 & 35.12  & 34.81 &  \underline{36.31}  & \textbf{36.56} \\
                             & SSIM$\uparrow$  & 0.4519   & 0.9659  & 0.9551  & 0.9591 & 0.9336 & 0.9614  & 0.9580  & 0.9652 & 0.9484  & 0.9512 & \underline{0.9611}  & \textbf{0.9748}  \\  
 \multirow{2}{*}{Mixed}      & PSNR$\uparrow$  & 26.59  & 32.86 & 32.47 & 33.83 & 36.53 & 30.20 & 34.26 & 33.44 & 34.69  & \underline{36.52} & 36.19  & \textbf{38.47} \\
                             & SSIM$\uparrow$  & 0.4757   & 0.7957  & 0.8002  &  0.8403  & 0.9189 & 0.7889  & 0.8584  & 0.7975  & 0.9084   & \underline{0.9196} & 0.9024   & \textbf{0.9329}  \\ \hline
\multicolumn{14}{c}{CVC09}    \\ \hline   
 \multirow{2}{*}{Gaussian}   & PSNR$\uparrow$  & 23.92  & 33.95 & 33.16 & 34.00 & 33.99  & 32.14 & 34.11 & {35.96} & 36.68   & 37.32 &  \underline{38.92}  & \textbf{40.02} \\
                             & SSIM$\uparrow$  & 0.3843   & 0.9574  & 0.9452  & 0.9647 & 0.9071 & 0.9624  & 0.9529  &  {0.9716}  & {0.9821}  & 0.9809 & \underline{0.9839}   & \textbf{0.9892}  \\
 \multirow{2}{*}{Uniform}    & PSNR$\uparrow$  & 26.80  & 36.57 & 35.54 & 35.63 & 37.74 & 32.24 & 38.72  &  {38.21} & 38.04  & 39.10 & \underline{40.60}  & \textbf{41.46} \\
                             & SSIM$\uparrow$  & 0.4976   & 0.9731  & 0.9653  & 0.9707 & 0.9480  & 0.9748  & 0.9740  &  0.9783  &  0.9789  & 0.9834 &  \underline{0.9854}   & \textbf{0.9907}  \\
 \multirow{2}{*}{Periodical} & PSNR$\uparrow$  & 25.96  & 35.06 & 33.91 & 32.90 & 35.06  & 30.53 & 35.41 & 34.90 & 35.32  & 34.87 &  \textbf{36.02}  & \underline{36.01} \\
                             & SSIM$\uparrow$  & 0.4099   &  {0.9761}  & 0.9652  & 0.9696 & 0.9596  & 0.9537  & 0.9411  &  {0.9761}  &  0.9771  & 0.9798 &  \underline{0.9844}   & \textbf{0.9881}  \\
 \multirow{2}{*}{Mixed}      & PSNR$\uparrow$  & 26.49  & 32.10 & 31.88  & 32.32  & 36.31 & 29.63 & 33.94 & 32.47  & 35.62  & \underline{36.88} & 35.44  & \textbf{38.48} \\
                             & SSIM$\uparrow$  & 0.4707   & 0.7543  & 0.7627  & 0.8020  & 0.9255 & 0.7535  & 0.8362  & 0.7549  & 0.9277   & \underline{0.9310} & 0.9022   & \textbf{0.9371}   \\ \hline
\end{tabular}
\vspace{-1mm} 
\end{table*}

\subsubsection{Self-Calibrated Branch}
As illustrated in Fig.~\ref{fig:4}(d), SCB is utilized to achieve powerful contextual information modeling and establish long-range dependencies at each spatial location~\cite{46}.
Given an input feature $\mathbf{X} \in \mathbb{R}^{C \times H \times W}$, the computation of the self-calibrated branch can be described as follows:
\begin{equation}
\mathbf{SC}={\delta_{s}}(\mathbf{X}+\mathcal{B}_{2}({conv}(\mathcal{A}_{2}(\mathbf{X})))),
\end{equation}
where $\mathcal{B}_{2}(\cdot)$ denotes a bilinear interpolation amplified by a factor of 2, $\mathcal{A}_{2}(\cdot)$ is an average pooling with filter size of 2$\times$2 and step size of 2.
After computing the self-calibrated matrix $\mathbf{SC} \in \mathbb{R}^{C \times H \times W}$, we can calculate the output of CSSC following Eq.~\ref{equ:6}. 
\vspace{-1mm}

\subsection{Loss Function}
Mean Square Error (MSE) loss is adopted to impose supervision on the training process. The loss function for $N$ image pairs can be expressed as follows:
\begin{equation}
\mathcal{L}=\frac{1}{N}\sum_{i=1}^{N}(\mathbf{I}_{O}^{i}-\mathbf{I}_{C}^{i})^{2},
\end{equation}
where $\mathbf{I}_{C}^{i}$ and $\mathbf{I}_{O}^{i}$  denote the $i$-th Ground Truth (GT) and the $i$-th output of ASCNet, respectively.

\section{EXPERIMENTAL RESULTS AND DISCUSSION}

\subsection{Experiment Settings}
We compare ASCNet to ten state-of-the-art infrared image destriping methods, including four canonical traditional methods (GF~\cite{1}, UTV~\cite{2}, LRSID~\cite{3} and SEID~\cite{SEID}), and six advanced deep learning-based methods (DMRN~\cite{5}, DLS-NUC~\cite{6}, SNRWDNN~\cite{7}, TSWEU~\cite{8}, Restormer~\cite{Restormer}, and DSCGAN~\cite{DSCGAN}). To guarantee an equitable comparison, the model-based methods use the default parameter provided by the authors. For learning-based methods that lack weights, we retrained them using the same datasets as our ASCNet and following their original parameter settings.

\subsubsection{Datasets}
To cover a wide range of scenarios, the experimental data used in this work is composed of the following three available IR datasets:
\begin{itemize}
    \item \textbf{ICSRN}~\cite{35}: Comprising 200 training images and 15 test images about IR clouds with a resolution of 579$\times$572. Taking into account the similarity within this training data, we selected 50 images from the ICSRN training set for training purposes, and the entire test set was utilized for testing.
    
    \item \textbf{INFRARED}~\cite{6}: He et al. released 100 high-quality 640$\times$480 infrared images containing rich daily life scenes. As this dataset was not pre-partitioned, we chose the first 70 images to construct our training dataset and reserved the remaining 30 for testing.
    
    \item \textbf{CVC09}~\cite{47}: Consisting of a large-scale sequence of far infrared (FIR) pedestrian images with a size of 640$\times$480 pixels. We selected 150 images showcasing diverse scenes from the training set for training, and 30 images from the test set were reserved for testing.
\end{itemize}

\begin{figure*}[t!]
    \centering
    \includegraphics[width=16cm]{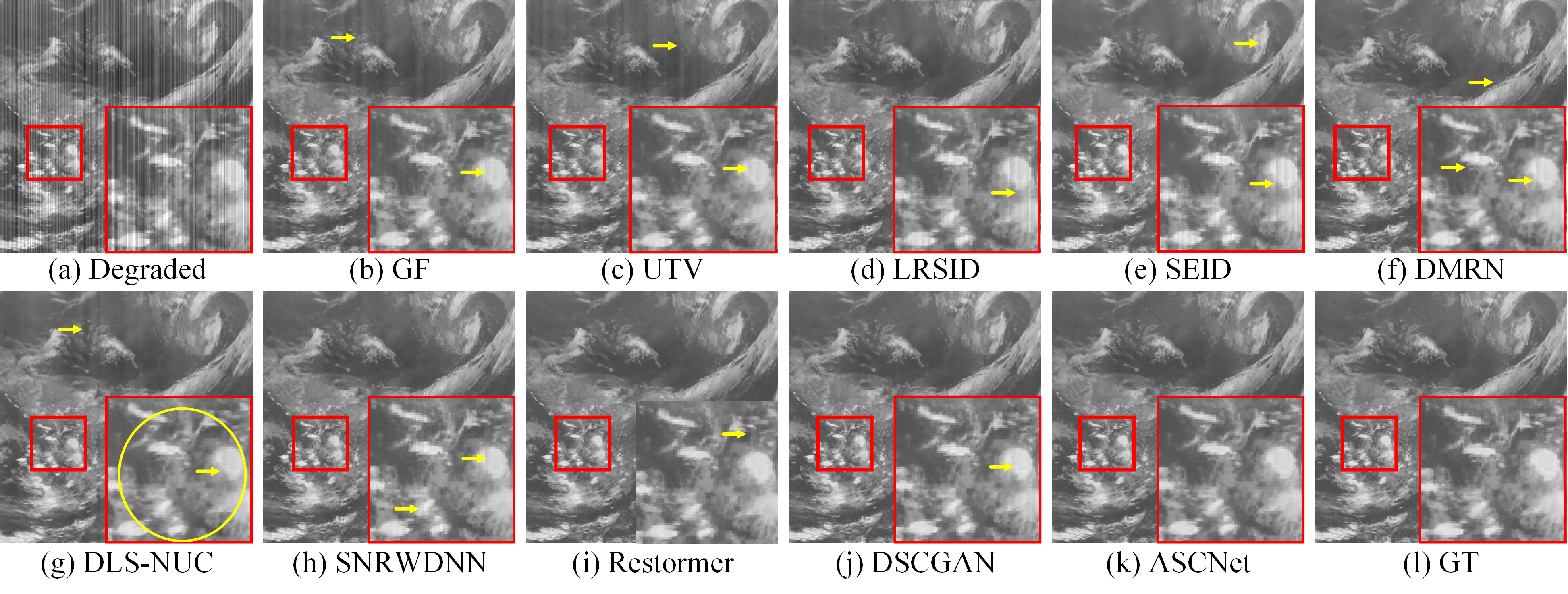}
    \vspace{-2mm}
    \caption{Image destriping results of different methods on ICSRN with Gaussian-distributed simulated noises.
    Yellow arrows and circles represent the stripe noise residue and blurred image details.
    }
    \vspace{-2mm}
    \label{fig:5}
\end{figure*}

\begin{figure*}[t!]
    \centering
    \includegraphics[width=17.2cm]{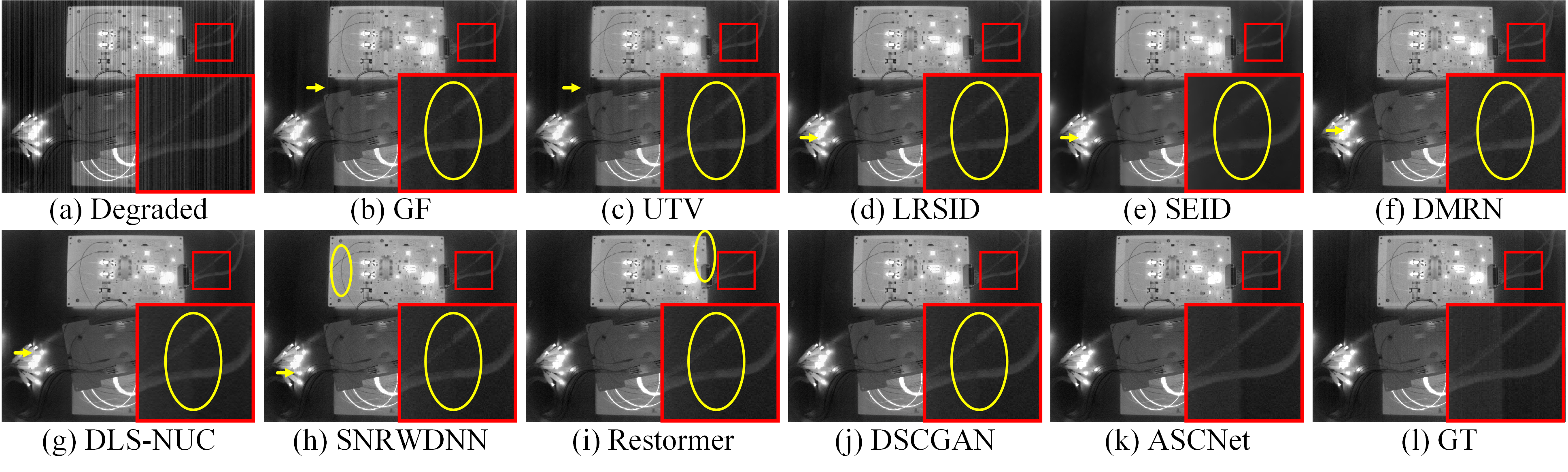}
    \vspace{-2mm}
    \caption{Image destriping results of different methods on INFRARED with uniform-distributed simulated noises. Stripe residuals and blurred image details are highlighted by yellow arrows and circles.
    }
    \vspace{-2mm}
    \label{fig:6}
\end{figure*}

\begin{figure*}[t!]
    \centering
    \includegraphics[width=17.2cm]{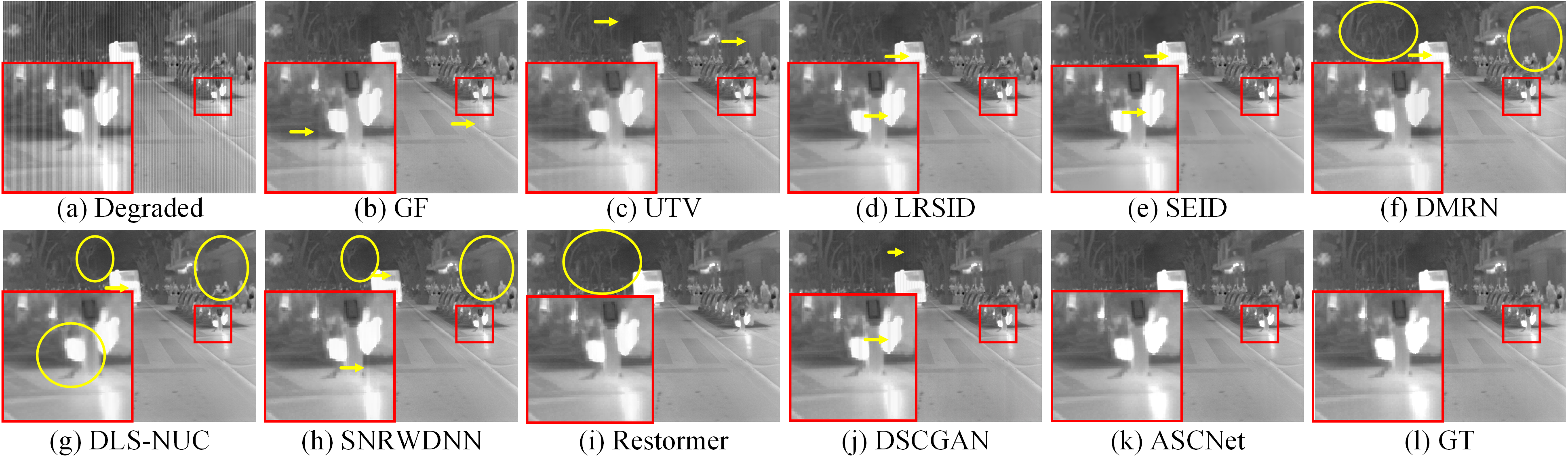}
    \vspace{-2mm}
    \caption{
    Image destriping results of different methods on CVC09 with periodical-distributed simulated noises. 
    Yellow arrows and circles represent the stripe noise residue and blurred image details. 
    }
    \vspace{-2mm}
    \label{fig:7}
\end{figure*}

\begin{figure*}[t!]
    \centering
    \includegraphics[width=17.2cm]{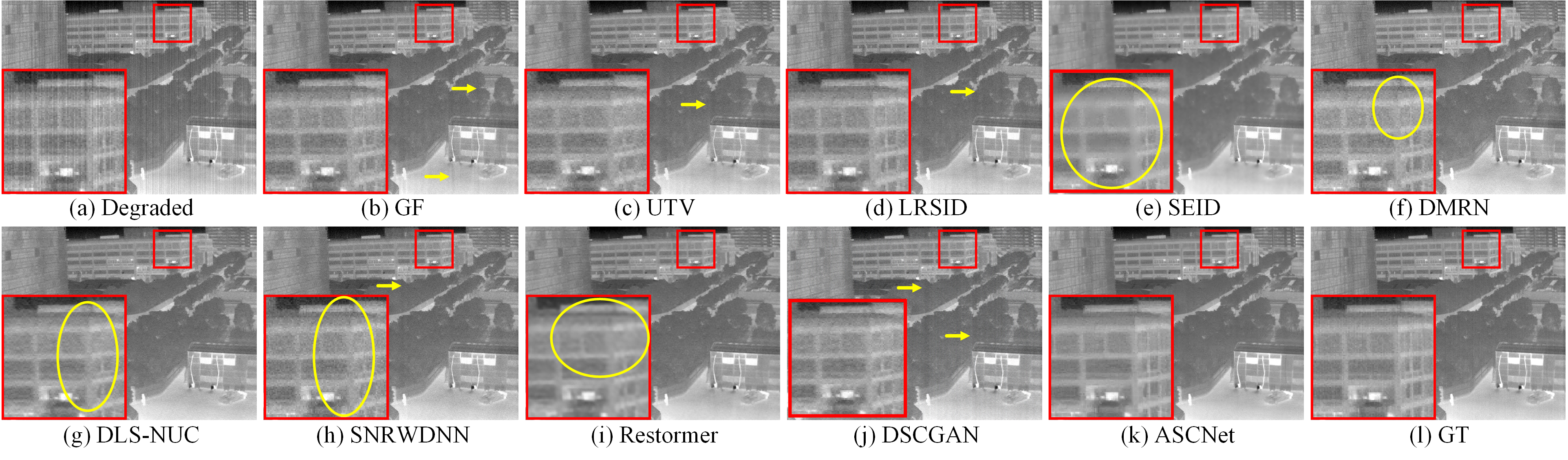}
    \vspace{-2mm}
    \caption{
    Image destriping results of different methods on INFRARED with Mixed simulated noises. 
    Stripe residuals and blurred image details are labeled by yellow arrows and circles.
    }
    \vspace{-3mm}
    \label{fig:add1}
\end{figure*}

\begin{figure*}[t!]
    \centering
    \includegraphics[width=17.2cm]{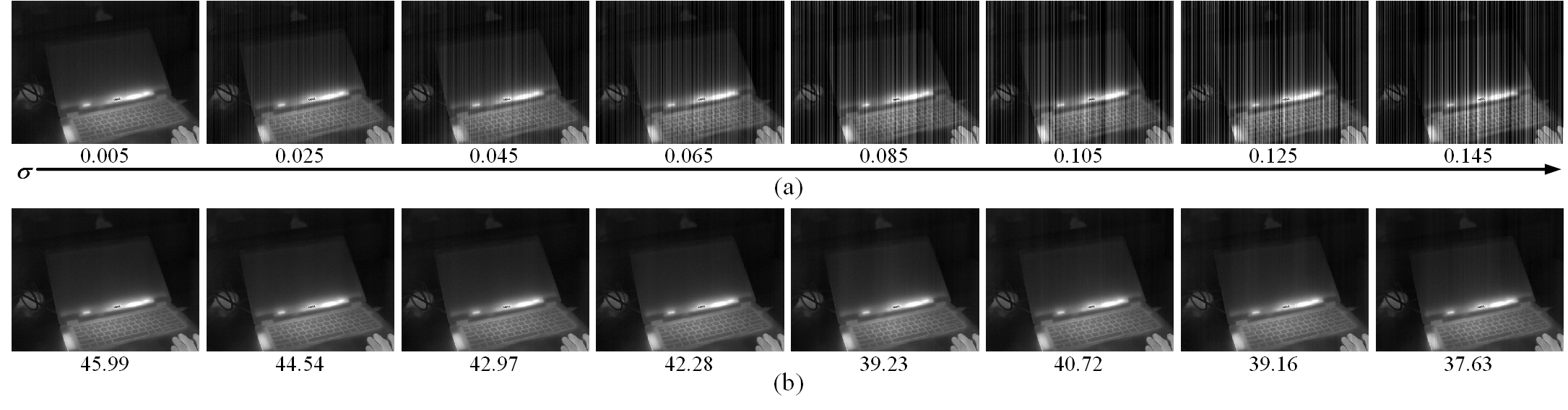}
    \vspace{-3mm}
    \caption{Exemplar stripy images and image destriping results of ASCNet. (a) Stripy image with different Gaussian-distributed noise levels. (b) Corresponding destriping results of ASCNet.
    }
    \vspace{-3mm}
    \label{fig:8}
\end{figure*}

Therefore, the dataset used in this study consists of 270 training images and 75 testing images.
Data augmentation techniques are applied to the training data, including rotation, scaling, and inversion~\cite{39},~\cite{Duan2},~\cite{kou3}. 
We then crop these augmented data using a sliding window with a patch size of 64×64 and a step size of 40×40, generating 180,486 patches.
The partitioned dataset and corresponding patch files can be obtained from \url{https://github.com/xdFai/ASCNet}.

\subsubsection{Noise Setting}
Considering that random noise often occurs together with stripe noise, we evaluated the proposed model's robustness using three typical distributions of stripe noise (Gauss, Uniform, and Periodical) and one type of mixed noise (random noise + stripe noise).
Previous studies have demonstrated the strong generalization ability of the cubic degradation model~\cite{6},~\cite{39}.
Therefore, we used cubic noise for the three types of stripe noise training. Specifically, the noise level of each order is set with a mean of 0, and the randomized standard deviation ranges from 0 to 0.15.
We added random noise with a mean of 0 and a standard deviation of 0.05 to the above stripe noise model for the mixed noise.

\subsubsection{Quantitative Metrics}
Following prior IR image destriping works~\cite{76},~\cite{3},~\cite{69}, we calculate the Peak Signal-to-Noise Ratio (PSNR) and Structural Similarity Index Measurement (SSIM) to evaluate the destriping performance. Higher PSNR and SSIM values indicate a closer similarity between the destriped and clean images.

\begin{table*}[t]
    \centering
    \caption{The \textit{$\rho$} and NIQE of different destriping methods on real stripe noises, the best and the second-best results are highlighted in bold and underlined}
    \vspace{-2mm}
    \renewcommand{\arraystretch}{1.2}
    \setlength\tabcolsep{2.8mm}
    \begin{tabular}{ccccccccccc}
    \hline \hline
    Metrics   & Degraded   & GF   & UTV    & LRSID     & DMRN    & DLS-NUC & SNRWDNN      & TSWEU  & DSCGAN   & ASCNet \\ \hline
    \textit{$\rho$}$\downarrow$  & 0.3108    &  0.1191 & 0.1311 & 0.1253 & 0.1208    & \textbf{0.0959}  & 0.1814  & 0.1389 & 0.1389 & \underline{0.1093}  \\
    NIQE$\downarrow$  & 62.164    & 8.7419 & 8.3217 & \underline{6.4368}  & 11.3561 & 8.5628  & 7.5688  &  6.9135 & 6.5936   & \textbf{5.9628} \\ \hline
    \end{tabular}
    \vspace{-2mm}
    \label{tab2}
\end{table*}

\begin{figure*}[t]
    \centering
    \includegraphics[width=17.2cm]{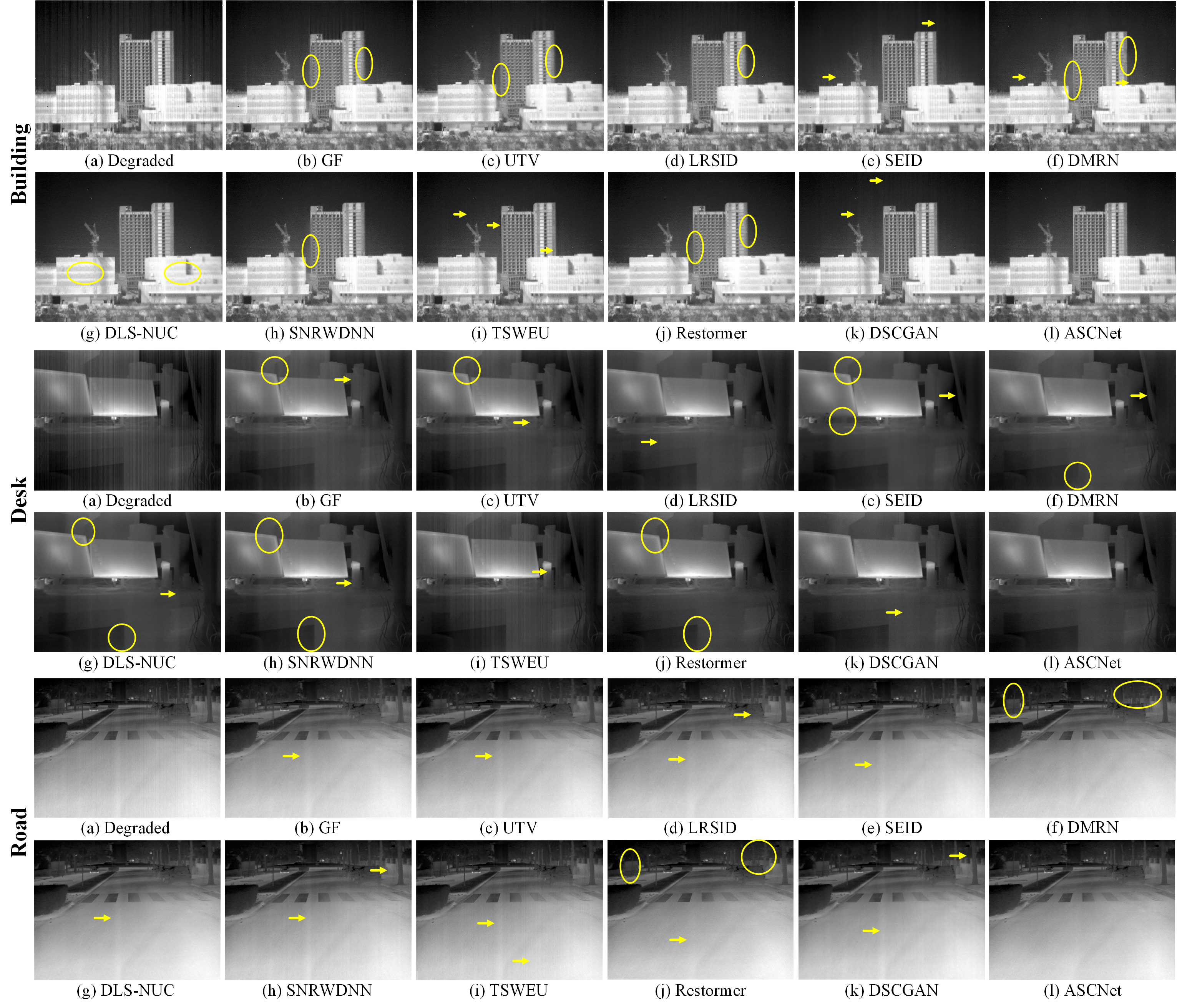}
    \vspace{-2mm}
    \caption{Destriping results for three sets of real-world IR stripe images. Zoom in to examine the details.}
    \vspace{-3mm}
    \label{fig:9}
\end{figure*}

\subsubsection{Implementation Details}
The basic width of ASCNet is set to 32, and the channel expansion factor $\eta$ in three RHDWTs is set to 2, 2, and 1 separately. We will discuss these hyperparameters in Section~\ref{Hyper-parameter}.
The batch size and epoch are set as 128 and 100, respectively. Our model was optimized using the Adam optimizer with an initial learning rate of 0.001 and gradually decreased to $1\times{{10}^{-6}}$ by the cosine annealing strategy. The number of warm-up epochs is set as 4. Meanwhile, we regularized the training by performing the orthogonalization technique~\cite{56} and applied the standard acceleration process to all wavelet transforms~\cite{51}. All experiments are conducted on a single Nvidia GeForce 3090 GPU, an Intel Core i7-12700KF CPU, and a 32 GB memory. 
The training process took approximately 15 hours.

\vspace{-2mm}
\subsection{Simulated Image Destriping}
\subsubsection{Quantitative Results}
We added the following three types of noises to each test set for validation:
\begin{itemize}
    \item \textbf{Gaussian}: Gaussian-distributed stripe noise with noise level {$\sigma\in[0, 0.10]$}.
    \item \textbf{Uniform}: Uniform-distributed stripe noise with noise level {$\mu\in[-0.10, 0.10]$}.
    \item \textbf{Periodical}: Randomly select a noise cycle period $T \in \left\{6,7,8,9\right\}$, and add Gaussian-distributed stripe noise with a noise level {$\sigma\in[0, 0.10]$} to period $T$.
    \item \textbf{Mixed}: Based on the \textbf{Gaussian}, further add Gaussian-distributed random noise with noise level {$\sigma\in[0, 0.05]$}.
\end{itemize}

The quantitative results are shown in Table~\ref{tab1}. The best and second-best results are indicated in bold and underlined. It is evident that our ASCNet outperforms the state-of-the-art methods. \emph{e.g.}, for the uniform-distributed stripe noise on INFRARED, ASCNet achieves PSNR values that are 0.43 dB and 3.01 dB higher than those of DSCGAN and Restormer, respectively.

\subsubsection{Visual Performance}
Qualitative destriping results are shown in Fig.~\ref{fig:5}, Fig.~\ref{fig:6}, Fig.~\ref{fig:7} and Fig.~\ref{fig:add1}.
Four sets of test images correspond to Gaussian, Uniform, Periodical and Mixed noises.
The effectiveness of ASCNet is apparent as it consistently delivers remarkable outcomes across various stripe scenarios. \emph{e.g.}, only our method successfully recovers the image details around the desk's edge in Fig.~\ref{fig:6}, while the other methods introduce artifacts and blurring.
For the mixed noise shown in Fig.~\ref{fig:add1}, our result is arguably clearer than the GT.

\subsubsection{Anti-noise Experiment}
Fig.~\ref{fig:8} showcases the visual results of ASCNet under various intensities of Gaussian-distributed stripe noise. Even with high noise intensity, ASCNet still exhibits excellent image restoration capabilities.

\begin{figure*}[t]
    \centering
    \includegraphics[width=17.2cm]{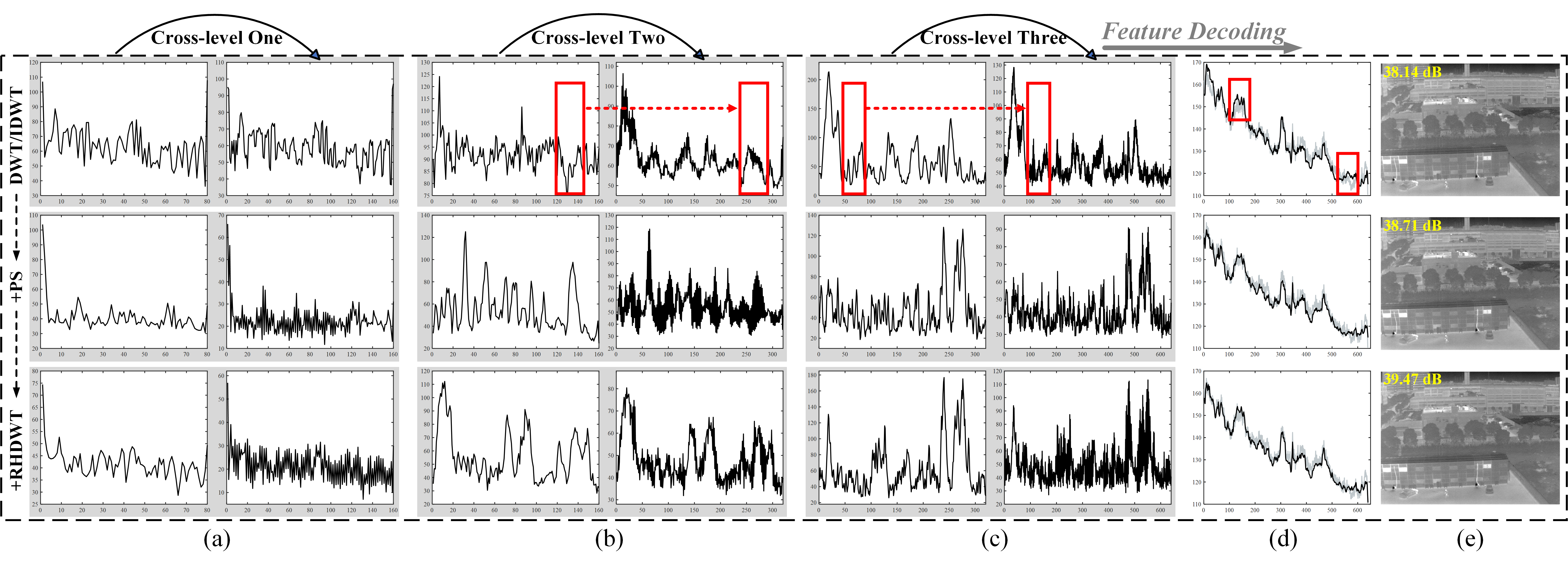}
    \vspace{-3mm}
    \caption{CMRCs of feature maps. (a), (b), and (c) denote three times upsampling for feature decoding. (d) showcases the CMRCs of the restored image and ground truth. (e) displays the visual result and the PSNR value. Red boxes highlight the column semantic gap caused by symmetric sampling.}
    \vspace{-1mm}
    \label{fig:10}
\end{figure*}

\subsection{Real Image Destriping}
To demonstrate the generalizability of ASCNet in varied real-world scenarios, experiments are conducted including 30 different scenes captured by the Thales Minie-D Long-Wave Infrared (LWIR) uncooled camera from CVC09 dataset, FLIR A65 LWIR detector from DLS-NUC, LWIR form MIRE and our InSb Mid-Wave Infrared (MWIR) cooled camera. These images contain rich and diverse stripe noise patterns.

\subsubsection{Quantitative Metrics}
Considering there are no reference images for real stripe images, following previous works~\cite{1},~\cite{27}, we employ roughness index \textit{$\rho$} and NIQE to quantitatively evaluate the performance of various destriping methods. 
As presented in Table~\ref{tab2}, ASCNet demonstrates outstanding performance, securing the best result on NIQE and the second-best on \textit{$\rho$}, underscoring its superior human visual perception.

\subsubsection{Visual Results}
Fig.~\ref{fig:9} illustrates the visual effects in three real-world scenes.
We can observe that the proposed method achieves optimal destriping results across various real noise patterns.
For example, only DMRN and ASCNet effectively suppress the concentrated noise in \textbf{``Road"}, but DMRN declines the localized contrast of the trees.
For more visual results see \url{https://github.com/xdFai/ASCNet}.

\begin{table}[t]
\centering
\caption{Based on U-Net, ablation study of the wavelet sampler (WS), DCR, asymmetric sampling (AS), RHDWT, and RCSSC module in average PSNR/SSIM on three test sets with Gaussian noise level {$\sigma\in[0,0.10]$}}
\label{tab3}
\vspace{-3mm}
\renewcommand{\arraystretch}{1.2}
\setlength\tabcolsep{1.3mm}
\begin{tabular}{ccccccc}\hline  \hline
 U-Net &+WS &+DCR  &+AS  & +RHDWT & +RCSSC & PSNR$\uparrow$/SSIM$\uparrow$      \\\hline
 \ding{51}    & \ding{55}    & \ding{55}    & \ding{55}     & \ding{55}  & \ding{55}    & 37.27/0.9823 \\\hline
 \ding{51}    & \ding{51}    & \ding{55}    & \ding{55}     & \ding{55}  & \ding{55}    & 37.68/0.9847 \\\hline
 \ding{51}     & \ding{51}    & \ding{51}     & \ding{55}     & \ding{55}  & \ding{55}    & 38.02/0.9854 \\\hline
 \ding{51}     & \ding{51}    & \ding{51}     & \ding{51}      & \ding{55}  & \ding{55}    & 38.34/0.9878 \\\hline
 \ding{51}     & \ding{51}    & \ding{51}     & \ding{51}      & \ding{51}   & \ding{55}    & 38.55/0.9885 \\\hline
 \ding{51}     & \ding{51}    & \ding{51}     & \ding{51}      & \ding{51}   & \ding{51}     & \textbf{40.11/0.9900} \\ \hline
\end{tabular}
\vspace{-2mm}
\end{table}

\begin{table}[t]
\centering
\vspace{-2mm}
\caption{The average PSNR/SSIM of the various symmetric sampling structures on three types of stripe noise}
\vspace{-2mm}
\label{tab4}
\renewcommand{\arraystretch}{1.2}
\setlength\tabcolsep{1.8mm}
\begin{tabular}{cccc}    \hline \hline
Method   & Gaussian     & Uniform      & Periodical    \\  \hline
DWT/IDWT~(Haar)      & \textbf{39.95}/0.9899 & \textbf{41.62/0.9920} & \textbf{36.57}/0.9846  \\
DWT/IDWT~(Db2)       & 39.88/0.9900 & 41.48/0.9918 & 36.72/0.9846   \\
DWT/IDWT~(Bior1.5)   & 39.76/\textbf{0.9901} & 41.27/0.9919 & 36.44/0.9843  \\
DWT/IDWT~(Sym5)      & \textbf{39.95}/0.9897 & 41.62/0.9917 & 36.07/\textbf{0.9850}   \\
SC/TC                & 39.75/0.9894 & 41.44/0.9913 & 36.32/0.9835 \\ 
PUS/PS               & 39.53/0.9899 & 41.20/0.9918 & 36.39/0.9840  \\ \hline
\vspace{-4mm}
\end{tabular}
\end{table}

\vspace{-3mm}
\subsection{Ablation Study and Further Discussion}
\label{sec: AbStudy}
Table~\ref{tab3} first demonstrates the effectiveness of each module in ASCNet. 
We can observe that the algorithm's performance progressively improves with the inclusion of the proposed structures.
Fig.~\ref{fig:10} further visualizes the effectiveness of PS and RHDWT for column modulation of stripe noises.
Next, we proceed with an in-depth analysis of ASCNet from three distinct perspectives.

\subsubsection{Asymmetric Sampling}
\label{sec: Asymmetric}
Based on ASCNet, Table~\ref{tab4} shows the quantitative results for the six symmetric samplers, including four typical discrete wavelet transforms: Haar, Daubechies2 (Db2), Biorthogonal1.5 (Bior1.5), and Symlets5 (Sym5), as well as Stride Convolution (SC)/Transpose Convolution (TC) and Pixel UnShuffle (PUS)/Pixel Shuffle (PS).
To ensure a fair comparison, we maintain the same parameters and computational complexity budget for all models.
It can be seen that the performance of wavelet transforms is generally superior to other structures. This is attributed to DWT's ability to aggregate noise features into distinct semantic patterns, facilitating effective neural network learning.
We also observe that the Haar wavelet performs slightly better than other wavelets. This is because the signal responses of adjacent columns in infrared FPA are independently and identically distributed, making the simple Haar wavelet better at highlighting the gradient changes of noise.
Subsequently, we use the HDWT as the downsampler and investigate the optimal upsampling structures, including Inverse Haar Discrete Wavelet Transform (IHDWT), Bilinear Interpolation (BI), Nearest Interpolation (NI), TC, and PS.
As shown in Table~\ref{tab5}, PS obtains the highest PSNR and SSIM values. This is because PS directly reconstructs the high-resolution feature based on the four neighboring semantic rearrangements with less semantic bias.

\begin{table}[t]
\centering
\caption{The average PSNR and SSIM of asymmetric sampling  on INFRARED and ICSRN with Gaussian noise level {$\sigma\in[0,0.10]$}}
\label{tab5}
\renewcommand{\arraystretch}{1.2}
\setlength\tabcolsep{1.6mm}
\begin{tabular}{cccccc}
\hline \hline
Method  & PSNR$\uparrow$ & SSIM$\uparrow$   & Params(M)$\downarrow$ & Flops(G)$\downarrow$ \\  \hline
IHDWT       & 39.18/40.05 & 0.9884/\textbf{0.9885} & 3.935    & \textbf{18.11}     \\
BI       & 39.55/39.95 & 0.9886/0.9884 & 3.930    & 20.08    \\
NI       & 38.89/39.93 & 0.9884/0.9879 & 3.930    & 20.08     \\
TC       & 39.52/40.06 & 0.9875/0.9871 & \textbf{3.923}   & 18.61         \\
PS       & \textbf{40.17/40.07} & \textbf{0.9887}/\textbf{0.9885} & 3.935    & \textbf{18.11}    \\ \hline
\vspace{-7mm}
\end{tabular}
\end{table}

\begin{table}[t]
\centering
\caption{The average PSNR and SSIM achieved by  
variants of RHDWT on three test sets with Gaussian noise level {$\sigma\in[0,0.10]$}}
\label{tab6}
\renewcommand{\arraystretch}{1.3}
\setlength\tabcolsep{2.4mm}
\begin{tabular}{ccccc}
 \hline  \hline
Method     & PSNR$\uparrow$   & SSIM$\uparrow$  & Params(M)$\downarrow$    & Flops(G)$\downarrow$  \\\hline
RHDWT~w~M       & 40.06   & 0.9897    & 3.935         & 18.11     \\
RHDWT~w~R       & 40.03    & 0.9891   &\textbf{3.216}   & \textbf{15.85}     \\ 
RHDWT~w~I       & 39.89   & 0.9895    & 4.710        & 19.92\\ 
RHDWT         & \textbf{40.11}   & \textbf{0.9900}   &  4.175  & 18.87\\   \hline
\end{tabular}
\vspace{-3mm}
\end{table}

\begin{figure}[t]
    \centering
    \includegraphics[width=0.40\textwidth]{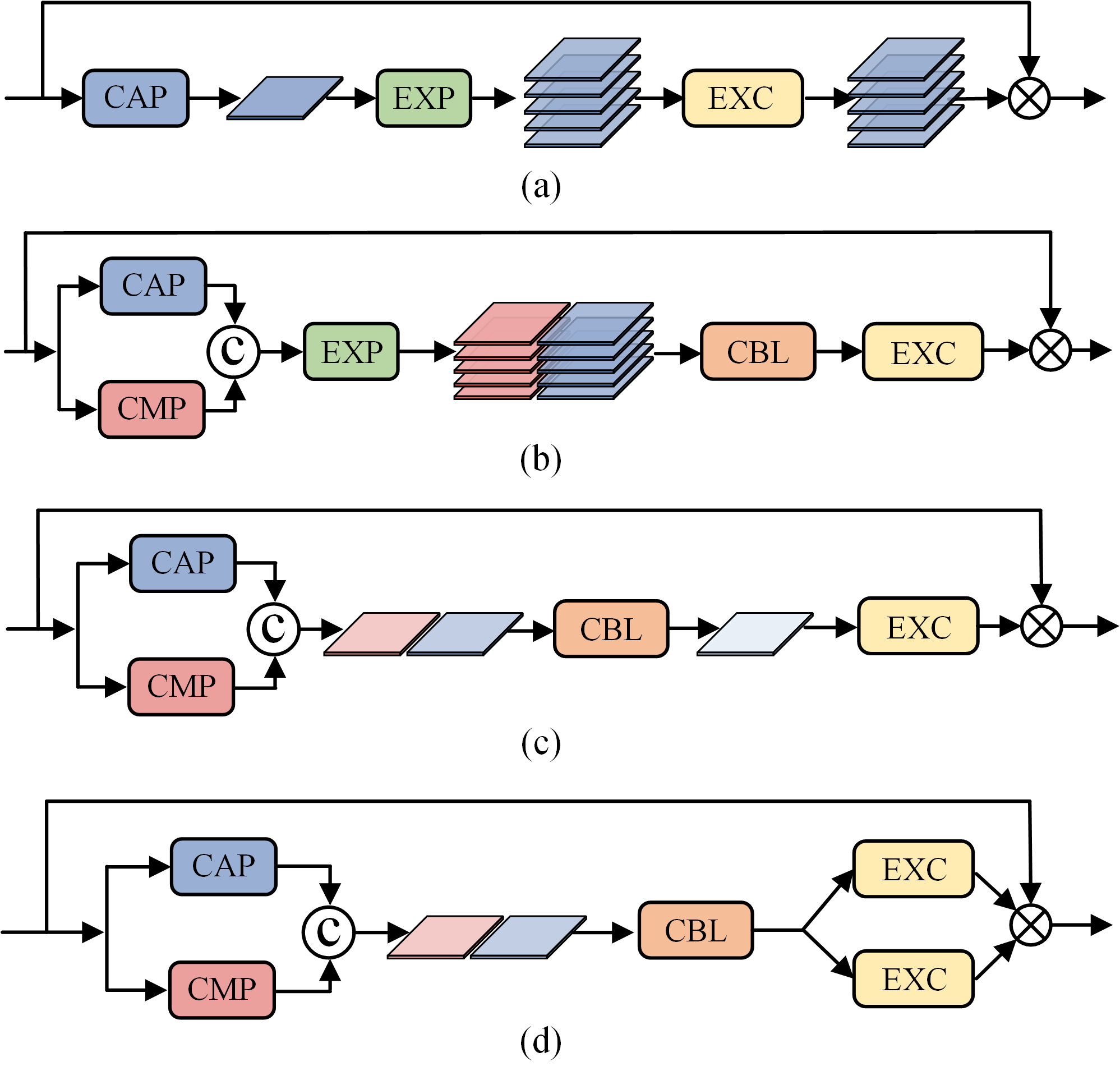}
    \vspace{-3mm}
    \caption{Different structures of the column attention mechanism. ``EXP” means expanding the feature size to match the input tensor, ``CBL" represents a set of Conv1×1+BN+LeakyReLU functions, and ``EXC" denotes the excitation module of the attention mechanism. (a) CCM~\cite{12}: traditional module in CSCNet.  (b) CAB\_v1: based on CCM, introduces dual-column-pooling. (c) CAB\_v2: based on CAB\_v1, introduces column-weight. (d) CAB: based on CAB\_v2, introduces dual-correction.}
    \vspace{-2mm}
    \label{fig:11}
\end{figure}

\begin{table}[t]
\centering
\vspace{-2mm}
\caption{The average PSNR and SSIM of the combinations of different branches under the CSSC architecture on three test sets with Gaussian noise {$\sigma\in[0,0.10]$}, the second to fifth rows represent different column calibration branches, the sixth and seventh rows denote the spatial attention and self-calibrated branches, respectively}
\vspace{-1mm}
\label{tab7}
\renewcommand{\arraystretch}{1.3}
\setlength\tabcolsep{1.2mm}
\begin{tabular}{c|c|c|c|c|c|c}
 \hline \hline
Method     & \textbf{K1}   & \textbf{K2}   & \textbf{K3}   & \textbf{K4}  & \textbf{K5} & \textbf{CSSC} \\ \hline
\multicolumn{1}{c|}{CCM~\cite{12}}      & \ding{55}        &  \ding{55}     & \ding{51}     & \ding{55}     &\ding{55}    &\ding{55}   \\
\multicolumn{1}{c|}{CAB\_v1} &  \ding{55}       & \ding{55}      &\ding{55}      & \ding{51}     & \ding{55}   & \ding{55}    \\
\multicolumn{1}{c|}{CAB\_v2} &  \ding{55}       & \ding{55}      &\ding{55}      & \ding{55}     & \ding{51}    & \ding{55}   \\
\multicolumn{1}{c|}{CAB}     & \ding{51}       & \ding{51}       & \ding{55}     & \ding{55}     & \ding{55}   & \ding{51}    \\ 
\multicolumn{1}{c|}{SCB~\cite{71}}       & \ding{51}       &  \ding{55}      & \ding{51}    & \ding{51}      & \ding{51}    & \ding{51}    \\
\multicolumn{1}{c|}{SAB~\cite{46}}       &  \ding{55}       & \ding{51}     & \ding{51}    & \ding{51}      & \ding{51}     & \ding{51}   \\  \hline
PSNR$\uparrow$   & 39.28 & 39.76 & 39.69 & 39.89 & 39.92   & \textbf{40.11} \\
SSIM$\uparrow$  & 0.9885  & 0.9888  & 0.9882  &  0.9891  & 0.9893  & \textbf{0.9900}  \\
Params(M)$\downarrow$  & \textbf{3.858}   & 3.915  & 4.054  & 4.112  & 4.174 &  4.175     \\
Flops(G)$\downarrow$   & \textbf{17.23} & 18.79 & 18.88  & 18.86  & 19.49 &  18.87 \\ \hline
\end{tabular}
\end{table}

\begin{figure}[t]
    \centering
    \includegraphics[width=0.48\textwidth]{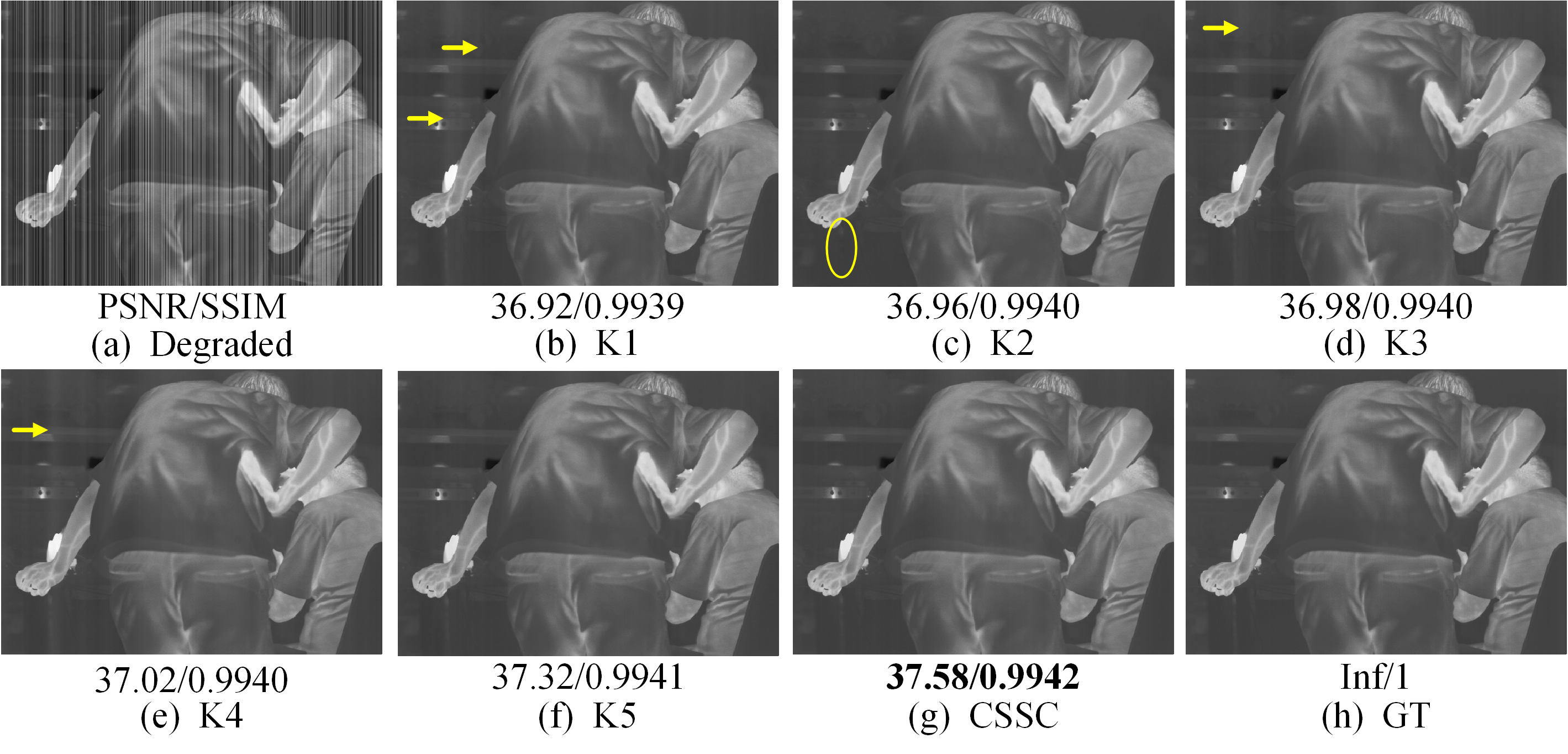}
    \vspace{-2mm}
    \caption{Qualitative destriping combinations of column attention branches on INFRARED with Gaussian noise {$\sigma\in[0,0.10]$}. Stripe residuals and blurred image details are highlighted by yellow arrows and circles.}
    \vspace{-2mm}
    \label{fig:XX}
\end{figure}

\subsubsection{Residual Haar Discrete Wavelet Transform}
\label{sec: RHDWT_Ablation}
To validate the effectiveness of double-branch sampling and explore the location of residual branch embedding, we built three structure variants: RHDWT with only model-driven branch ($ RHDWT~w~M$), RHDWT with only residual branch ($ RHDWT~w~R$), RHDWT with embedding residual branch into model-driven branch immediately after wavelet sub-band concatenate ($ RHDWT~w~I$), respectively.
The quantitative results are exhibited in Table~\ref{tab6}.
Comparing $RHDWT~w~M$ and $RHDWT~w~R$, despite $RHDWT~w~M$ consuming more parameters and computational complexity, it effectively improves PSNR and SSIM values by combining the directional prior of the stripe noise.
Our RHDWT achieves the best results with acceptable parameters.
This illustrates that paralleling stripe-directional prior knowledge with data-driven aggregating semantics can enhance the effectiveness of features.


\subsubsection{Column Spatial Self-Correction}
As shown in Fig.~\ref{fig:11}, we showcase three different column attention branches to demonstrate the effectiveness of the proposed CSSC.
Table~\ref{tab7} shows the results of various column branches combined with spatial and self-calibration branches ($K1-K5,~CSSC$) based on our ASCNet.

\begin{itemize}
    \item ${K1, K2, CSSC:}$ It demonstrates that both self-calibration and spatial attention branches are essential. Without using the spatial attention or self-calibration branch, the PSNR will decrease by 0.83 dB and 0.34 dB, respectively.
    \item ${K3, K4:}$ Despite a larger model size of 0.058 M, by utilizing the dual-column-pooling, CAB\_v1 achieved 0.30 dB higher PSNR and 0.0009 higher SSIM than CCM.
    \item ${K4, K5, CSSC:}$ Using column-weigh and dual-correction strategies, CSSC improves the value of PSNR by 0.03 dB and 0.19 dB without parameters increasing. 
\end{itemize}

We further compare the visual performance of different combinations of column attention branches in Fig.~\ref{fig:XX}. 
The results demonstrate that the strategies employed in CSSC exhibit more successful stripe noise reduction.

\begin{table}[t]
\centering
\caption{Hyperparameter study of the model in average PSNR and SSIM on three test sets with Gaussian noises level {$\sigma\in[0,0.10]$}}
\vspace{-2mm}
\label{tab10}
\renewcommand{\arraystretch}{1.2}
\setlength\tabcolsep{2.2mm}
\begin{tabular}{ccccc} \hline  \hline
Hyper-param & PSNR$\uparrow$ & SSIM$\uparrow$ & Params(M)$\downarrow$ & Flops(G)$\downarrow$ \\ \hline 
\multicolumn{5}{c}{The channel expansion factor of the third RHDWT} \\\hline
$\boldsymbol{\eta_{3} = 1}$  & 40.11  & 0.9900  & 4.175   &  18.87           \\
$\eta_{3}$ = 2        & 40.10  & 0.9901  & 7.341   &  21.79        \\\hline
\multicolumn{5}{c}{The   loss function of the model}       \\\hline
MAE loss         & 39.99 &  0.9898 & 4.175  &  18.87       \\
\textbf{MSE} \textbf{loss}   & 40.11 &  0.9900 & 4.175  &  18.87          \\\hline
\multicolumn{5}{c}{The   basic width of the model}         \\\hline
W = 8         & 39.43  & 0.9874 & 0.264  &  1.186        \\
W = 16        & 39.82  & 0.9876 & 1.046   &  4.725       \\
\textbf{W = 32} & 40.11  & 0.9900 & 4.175   &   18.87       \\
W = 48        & 40.18  & 0.9908 & 16.68  &  75.42     \\ \hline
\end{tabular}
\end{table}

\begin{figure}[t]
    \centering
    \includegraphics[width=8.5cm]{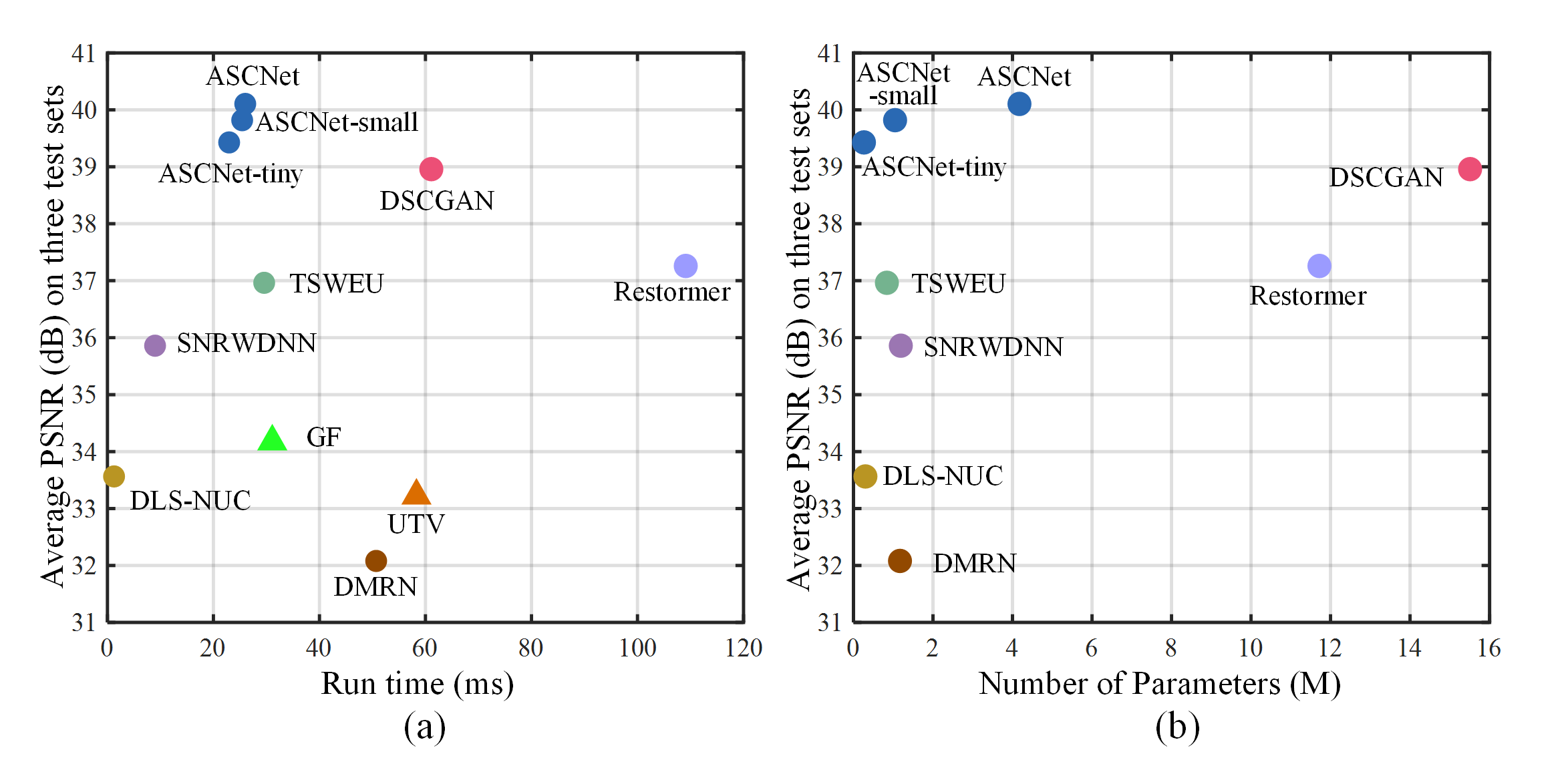}
    \vspace{-3mm}
    \caption{(a) Relationship among run time (ms) and PSNR. (b) Relationship between the number of parameters (M) and PSNR.}
    \label{fig:12}
\end{figure}

\subsection{Hyperparameter Analysis and the Efficiency of ASCNet}
\label{Hyper-parameter}
Since ASCNet performs many computations in the RCSSC to obtain remote fine-grained contextual information, the channel expansion factor $\eta$ of the third RHDWT, here named $\eta_{3}$, is set as 1 to mitigate the parameters burden. 
In this section, we validate the hyperparameters of ASCNet from $\eta$ and the base width of the mode $W$.
As shown in Table~\ref{tab10}, the structure with $\eta_{3}$ = 2 has similar performance as $\eta_{3}$ = 1 but brings larger parameters and computational complexity, so ASCNet sets $\eta_{3}$ to 1.
We also observe that MSE loss obtains better results than Mean Absolute Error (MAE) loss.
Additionally, the number of parameters and flops, as well as the destriping performance, show significant improvement as the base width of the model gradually changes from 8 to 32.
However, the model with $W$ = 48 quadruples the number of parameters with little performance gain.
Therefore, we set the basic width of ASCNet as 32 and defined ASCNet with $W$ = 18 and $W$ = 6 as \textit{ASCNet-small} and \textit{ASCNet-tiny}.
Model efficiency is shown in Fig.~\ref{fig:12}, GF, UTV, LRSID, DMRN, DLS-NUC, and TSWEU ran on the MATLAB 2020b platform, SNRWDNN ran on the TensorFlow 2.2.0 framework, Restormer, DSCGAN and ASCNet conduct on Pytorhc 1.7.1. Our ASCNet and its variants have an acceptable running speed and optimal performance.

\begin{table}[t]
\centering
\caption{Detection results on real stripe images and images processed by ASCNet in average ${P}_{d}(\%)$, ${F}_{a}({10}^{-6})$, $IoU(\%)$, $F$-$measure(\%)$}
\label{tab:8}
\renewcommand{\arraystretch}{1.2}
\setlength\tabcolsep{2mm}
\begin{tabular}{cccc}  \hline  \hline
\multirow{2}{*}{Method} & ACM~\cite{48} & RDIAN~\cite{78} & DNA-Net~\cite{49}  \\ \cline{2-4}
                         & Dagred/ASCNet & Dagred/ASCNet & Dagred/ASCNet  \\ \hline
${P}_{d}$$\uparrow$             & 77.97/\textbf{80.00}   & 78.73/\textbf{81.51}   & 86.46/\textbf{89.37}   \\
${F}_{a}$$\downarrow$          & 1.648/\textbf{1.631}    & 0.456/\textbf{0.435}   & 1.381/\textbf{1.236}   \\
IOU$\uparrow$            & 45.42/\textbf{46.13}   & 64.60/\textbf{65.06}   & 65.88/\textbf{68.11}   \\
F-measure$\uparrow$       & 62.47/\textbf{63.13}   & 78.50/\textbf{78.83}   & 79.43/\textbf{81.03} \\  \hline
\end{tabular}
\end{table}

\begin{figure}[t!]
    \centering
    \includegraphics[width=8.6cm]{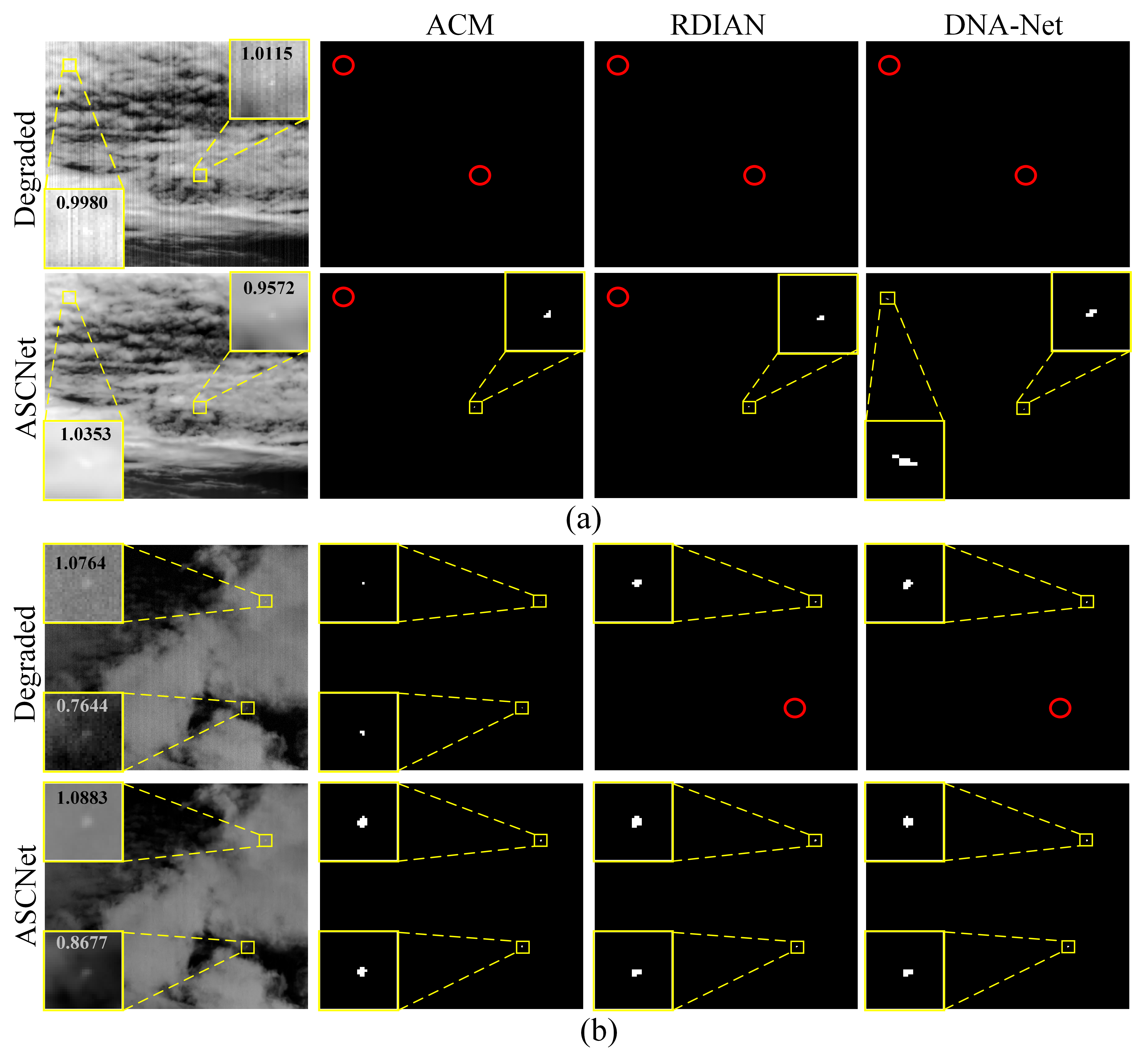}
    \vspace{-1mm}
    \caption{(a) IRSTD results on simulated mixed noise. (b) IRSTD results on real stripe noise mixed with simulated Gaussian noise. Yellow boxes and red circles represent correctly detected targets and miss detection, respectively.}
    \label{fig:14}
    \vspace{-3mm}
\end{figure}

\subsection{Task-driven Evaluate: Infrared Small Target Detection}
Taking into account that destriping can be treated as a pre-processing for subsequent applications, we showcase ASCNet's efficacy in the IRSTD task. 
Firstly, we conduct 100 IR images containing three unmanned aerial vehicles with stripe noises acquired by long-wave uncooled cameras.
Three SOTA learning-based IRSTD methods, ACM~\cite{48}, RDIAN~\cite{78}, and DNA-Net~\cite{49}, are selected to evaluate the advantage that ASCNet brings to the IRSTD task.
In table~\ref{tab:8}, Probability of Detection (${{P}_{d}}$), False-Alarm Rate (${{F}_{a}}$), Intersection over Union (IoU) and F-measure are utilized to evaluate the detection result~\cite{72}.
It can be seen that the performance of all three detectors is improved after our ASCNet processing.
Fig.~\ref{fig:14} shows the two scenes' visual effects of stripe removal and target detection.
The targets in the degraded images exhibit a low Signal-to-Noise Ratio (SNR) (with the neighbourhood set to 10) and are contaminated by stripe and random noise.
Our ASCNet effectively restores the target and its local region, enhancing the SNR of most targets. Consequently, many previously missed targets are successfully detected, and the outlines of the targets are more accurately delineated.
These results indicate the effectiveness of ASCNet on IRSTD.

\section{CONCLUSION}
In this paper, we propose a novel asymmetric sampling correction network for IR image destriping. 
Based on the observation of semantic crosstalk with stripe noise, we introduce pixel shuffle as an upsampler to prevent excessive decoding of apriori features and promote cross-level semantic articulation. 
We also propose the residual Haar discrete wavelet transform as a downsampler to incorporate stripe-directional prior knowledge and data-driven semantic interaction, enhancing the presentation of image features.
Moreover, a column non-uniformity correction module is designed to enhance remote dependence of column characteristics in the global context.
Extensive experiments demonstrate that the proposed method performs well against the state-of-the-art methods in both synthetic and real settings. We also confirm its effectiveness in enhancing the precision of the downstream IRSTD task.

\bibliographystyle{IEEEtran}
\small\bibliography{IEEEabrv,reference}

\begin{IEEEbiography}[{\includegraphics[width=1in, height=1.4in, clip, keepaspectratio]{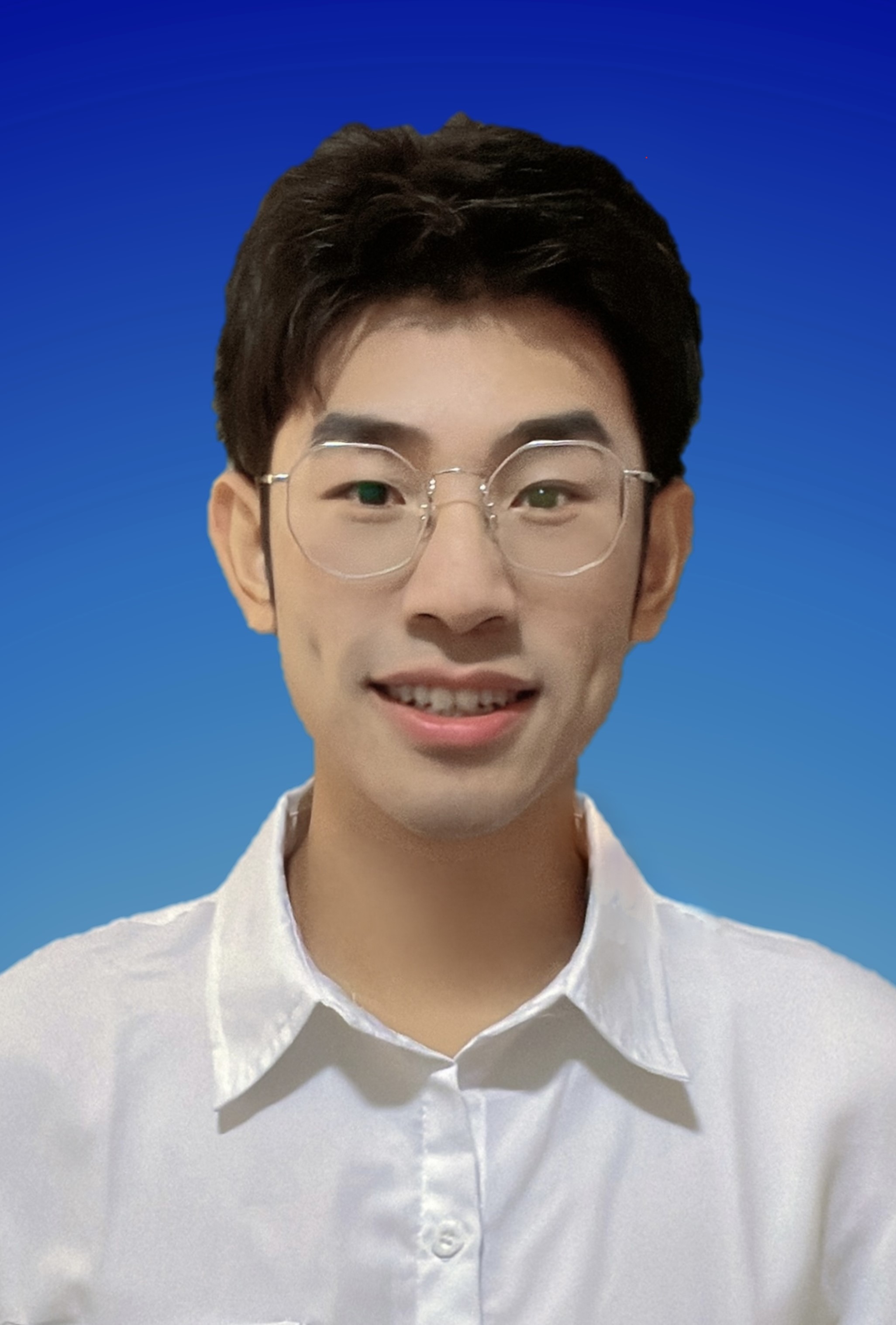}}]{Shuai Yuan}
received the B.S. degree from Xi'an Technological University, Xi'an, China, in 2019. He is currently pursuing a Ph.D. degree at Xidian University, Xi’an, China. He is currently studying at the University of Melbourne as a visiting student, working closely with Dr. Naveed Akhtar. His research interests include infrared image understanding, remote sensing, and deep learning.
\end{IEEEbiography}

\begin{IEEEbiography}[{\includegraphics[width=1in, height=1.3in, clip, keepaspectratio]{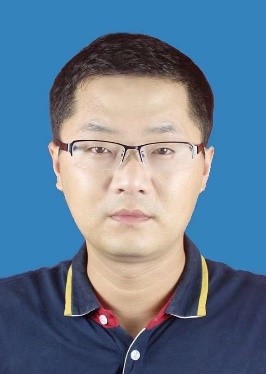}}]{Hanlin Qin}
received the B.S and Ph.D. degrees from Xidian University, Xi'an, China, in 2004 and 2010. He is currently a full professor at the School of Optoelectronic Engineering, Xidian University. He authored or co-authored more than 100 scientific articles. His research interests include target detection, human-computer interaction, and photonic chip.
\end{IEEEbiography}

 \begin{IEEEbiography}[{\includegraphics[width=1.5in, height=1.4in, clip, keepaspectratio]{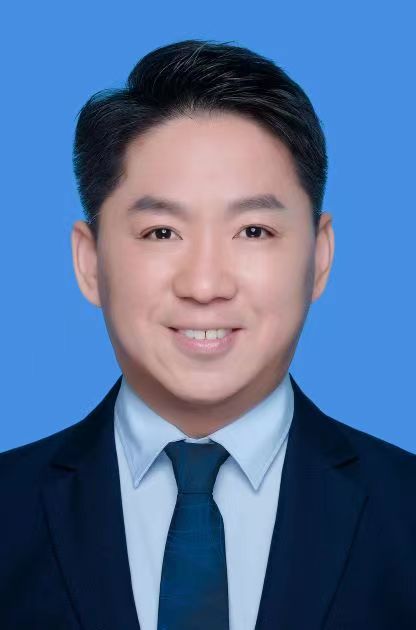}}]{Xiang Yan} received the B.S and Ph.D. degrees from Xidian University, Xi'an, China, in 2012 and 2018. He was a visiting Ph.D. Student with the School of Computer Science and Software Engineering, Australia, from 2016 to 2018, working closely with Prof. Ajmal Mian. He is currently an associate professor at Xidian University, Xi'an, China. His current research interests include image processing, computer vision and deep learning.
\end{IEEEbiography}

\begin{IEEEbiography}[{\includegraphics[width=1.3in,height=1.3in,clip,keepaspectratio]{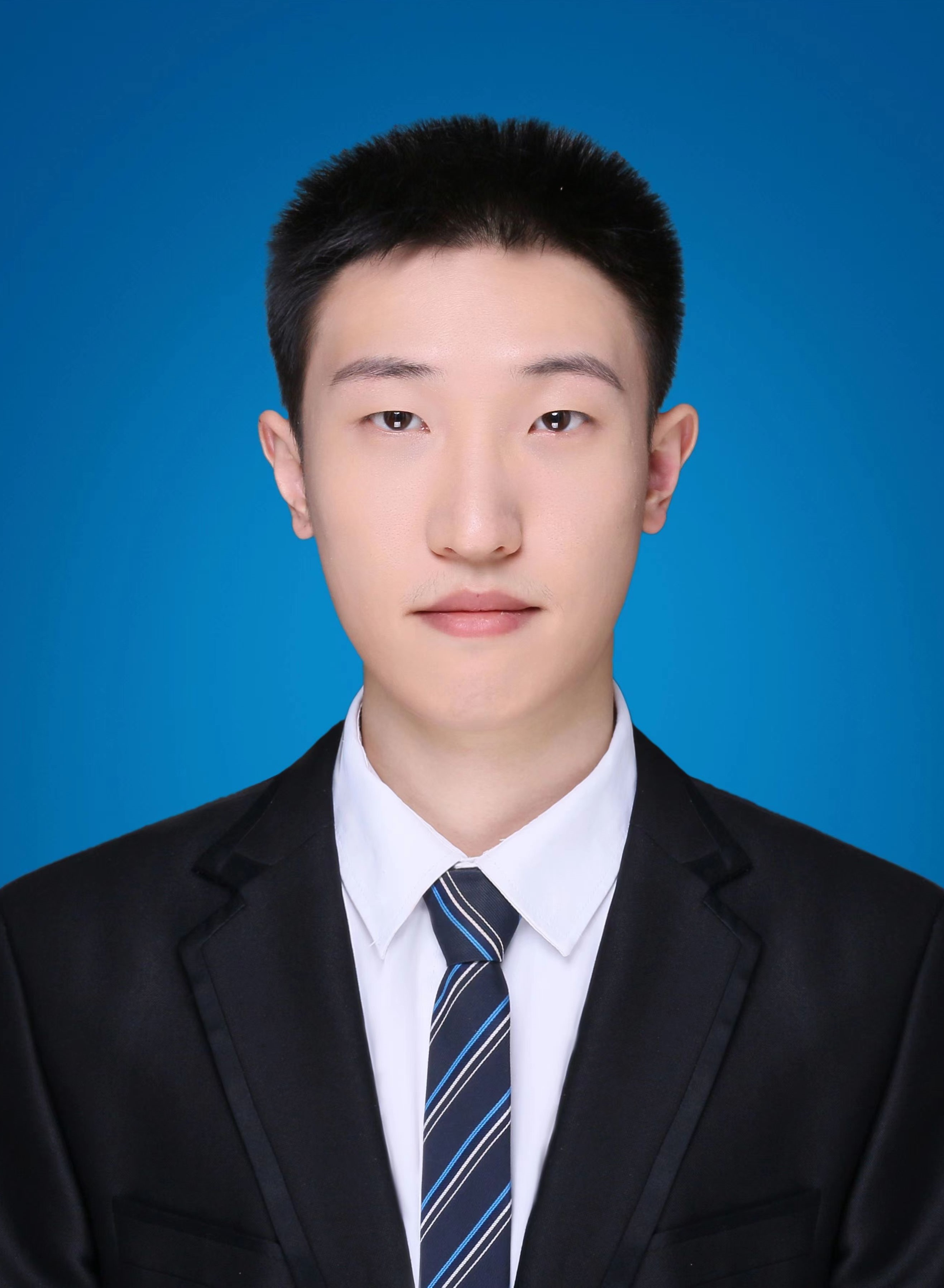}}]{Shiqi Yang}
received the B.S. degree in electronic science and technology from Xidian University, Shanxi, China, in 2022. He is currently pursuing the M.S. degree with Xidian University. His current research interests include deep learning, image destriping, and infrared small target detection.
\end{IEEEbiography}

\begin{IEEEbiography}
[{\includegraphics[width=1.3in,height=1.3in, clip,keepaspectratio]{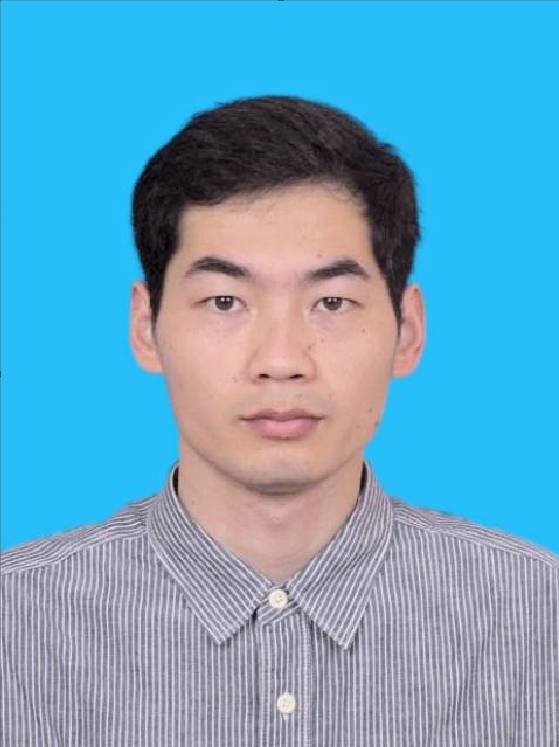}}]{Shuowen Yang} 
received the B.S. degree in electronic science and technology in 2016 and the Ph.D. degree in physical electronics in 2023, both from Xidian University. During pursuing the Ph.D. degree, he studied in the University of Granada from 2021 to 2022, working closely with Prof. Rafael Molina. 
\end{IEEEbiography}
\vfill

\begin{IEEEbiography}[{\includegraphics[width=1in, height=1.2in, clip, keepaspectratio]{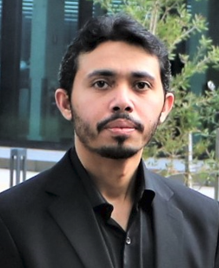}}]{Naveed Akhtar} is a Senior Lecturer  at the University of Melbourne. He received his PhD in Computer Science from the University of Western Australia and Master degree from Hochschule Bonn-Rhein-Sieg, Germany. He is a recipient of the Discovery Early Career Researcher Award from the Australian Research Council. He is a Universal Scientific Education and Research Network Laureate in Formal Sciences, and a recipient of Google Research Scholar Program award in 2023. He was a finalist of the Western Australia's Early Career Scientist of the Year 2021.  He has served as an ACM Distinguished Speaker (2021-2024) and serves as an Associate Editor of IEEE Trans. Neural Networks and Learning Systems. He has also served as an Area Chair for reputed conferences like IEEE Conf. on Computer Vision and Pattern Recognition (CVPR) and European Conference on Computer Vision (ECCV) on multiple occasions.
\end{IEEEbiography}

\begin{IEEEbiography}[{\includegraphics[width=1.1in,height=1.2in, clip,keepaspectratio]{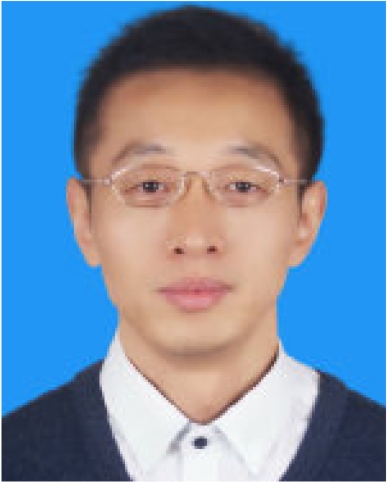}}]{Huixin Zhou}  (Member, IEEE) received the B.S.,
M.S., and Ph.D. degrees in optical engineering from
Xidian University, Xi’an, China, in 1996, 2002, and
2004, respectively.

He is currently the Vice-Dean and a Professor with the School of Physics, Xidian University. His research interests include remote sensing image processing, photoelectric imaging, real-time image processing, and target detection and tracking.

Dr. Zhou is also a Standing Member of the Optoelectronic Technology Professional Committee of the Chinese Society of Astronautics, a Senior Member of the Chinese Optical
Society, and a member of the Optical Society of America and the Shaanxi Provincial Physical Society. He is an Associate Editor of the journal of \textit{Infrared Physics} \& \textit{Technology}.
\end{IEEEbiography}

\end{document}